\documentclass[]{fairmeta}

\usepackage[utf8]{inputenc}
\usepackage[T1]{fontenc}
\usepackage{url}
\usepackage{booktabs}
\usepackage{amsmath}
\usepackage{amsfonts}
\usepackage{amssymb}
\usepackage{nicefrac}
\usepackage{microtype}
\usepackage{xcolor}
\usepackage{graphicx}
\usepackage{subcaption}
\usepackage[most]{tcolorbox}
\usepackage{tikz}
\usetikzlibrary{positioning,arrows.meta,calc,fit,backgrounds,shapes.geometric,shapes.misc}
\usepackage{listings}
\usepackage{tabularx}
\usepackage{array}
\usepackage{enumitem}
\usepackage{float}
\usepackage{pgf}
\usepackage{colortbl}
\usepackage{rotating}
\usepackage{longtable}
\usepackage[english]{babel}
\usepackage{csquotes}

\definecolor{bluelink}{RGB}{0,113,188}
\definecolor{citecolor}{HTML}{0071bc}
\hypersetup{
    colorlinks=true,
    citecolor=citecolor,
    filecolor=citecolor,
    linkcolor=citecolor,
    urlcolor=citecolor
}

\lstdefinelanguage{mtssjson}{
  morestring=[b]",
  morecomment=[l]{//},
  literate=
    *{0}{{{\color{purple!70!black}0}}}{1}
     {1}{{{\color{purple!70!black}1}}}{1}
     {2}{{{\color{purple!70!black}2}}}{1}
     {3}{{{\color{purple!70!black}3}}}{1}
     {4}{{{\color{purple!70!black}4}}}{1}
     {.}{{{\color{purple!70!black}.}}}{1}
     {:}{{{\color{black!70}:}}}{1}
     {,}{{{\color{black!50},}}}{1},
  stringstyle=\color{teal!60!black},
  commentstyle=\color{gray!60!black}\itshape,
  showstringspaces=false,
}
\newcounter{promptbox}
\renewcommand{\thepromptbox}{\Alph{promptbox}}
\tcbset{
  promptbox/.style={
    enhanced, breakable,
    colback=gray!4, colframe=gray!55!black,
    coltitle=white, colbacktitle=gray!55!black,
    fonttitle=\bfseries\small,
    boxrule=0.4pt, arc=2pt,
    left=6pt,right=6pt,top=4pt,bottom=4pt,
    before skip=8pt,after skip=8pt,
    fontupper=\small\ttfamily\raggedright,
  },
}
\newcommand{\promptlabel}[1]{\refstepcounter{promptbox}\thepromptbox\label{#1}}

\newlength\savewidth
\newcolumntype{x}[1]{>{\centering\arraybackslash}p{#1pt}}
\newcolumntype{y}[1]{>{\raggedright\arraybackslash}p{#1pt}}
\newcolumntype{z}[1]{>{\raggedleft\arraybackslash}p{#1pt}}

\renewcommand{\paragraph}[1]{\vspace{1.25mm}\noindent\textbf{#1}}

\title{SARA: Semantically Adaptive Relational \\[3mm] Alignment for Video Diffusion Models}

\renewcommand\authorlist{%
  \authorformat[1,2\dagger]{Jiesong Lian}, \authorformat[2]{Zixiang Zhou}, \authorformat[3]{Ruizhe Zhong}, \authorformat[2\ddagger]{Yuan Zhou}, \authorformat[2]{Qinglin Lu},\\
  \authorformat[1\text{\S}]{Rui Wang}, \authorformat[1]{Long Hu}, \authorformat[1]{Yixue Hao}, \authorformat[4]{Baoru Huang}%
}

\renewcommand\affiliationlist{%
  \affiliationformat[1]{Huazhong University of Science and Technology}, \affiliationformat[2]{Tencent Hunyuan},\\
  \affiliationformat[3]{Shanghai Jiao Tong University}, \affiliationformat[4]{University of Liverpool}%
}

\newcommand{\papertitle}{SARA: Semantically Adaptive Relational Alignment for Video Diffusion Models}

\usepackage{fancyhdr}
\fancypagestyle{titlepagewithlogo}{
  \fancyhf{}
  
  \rhead{\includegraphics[height=0.6cm]{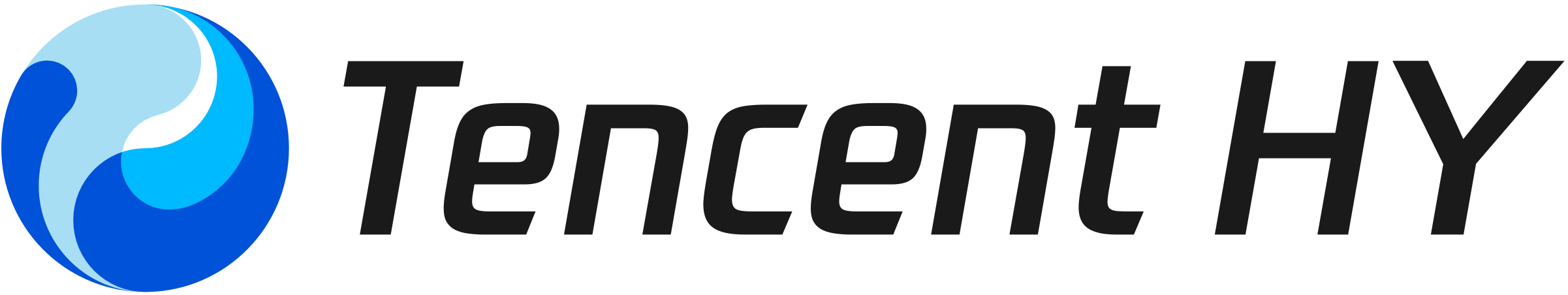}}
}

\abstract{}

\begin{document}
\maketitle
\thispagestyle{titlepagewithlogo}
\vspace{-10mm}

\renewcommand{\thefootnote}{\fnsymbol{footnote}}
\begingroup
\makeatletter
\long\def\@makefntext#1{\noindent#1}
\makeatother
\footnotetext{%
  $^\dagger$Work done during internship at Tencent Hunyuan.\\
  $^\ddagger$Project leader.\\
  $^{\text{\S}}$Corresponding author. {\tt ruiwang2020@hust.edu.cn}}
\endgroup
\renewcommand{\thefootnote}{\arabic{footnote}}

\vspace{2mm}
\begin{center}
  \includegraphics[width=\linewidth]{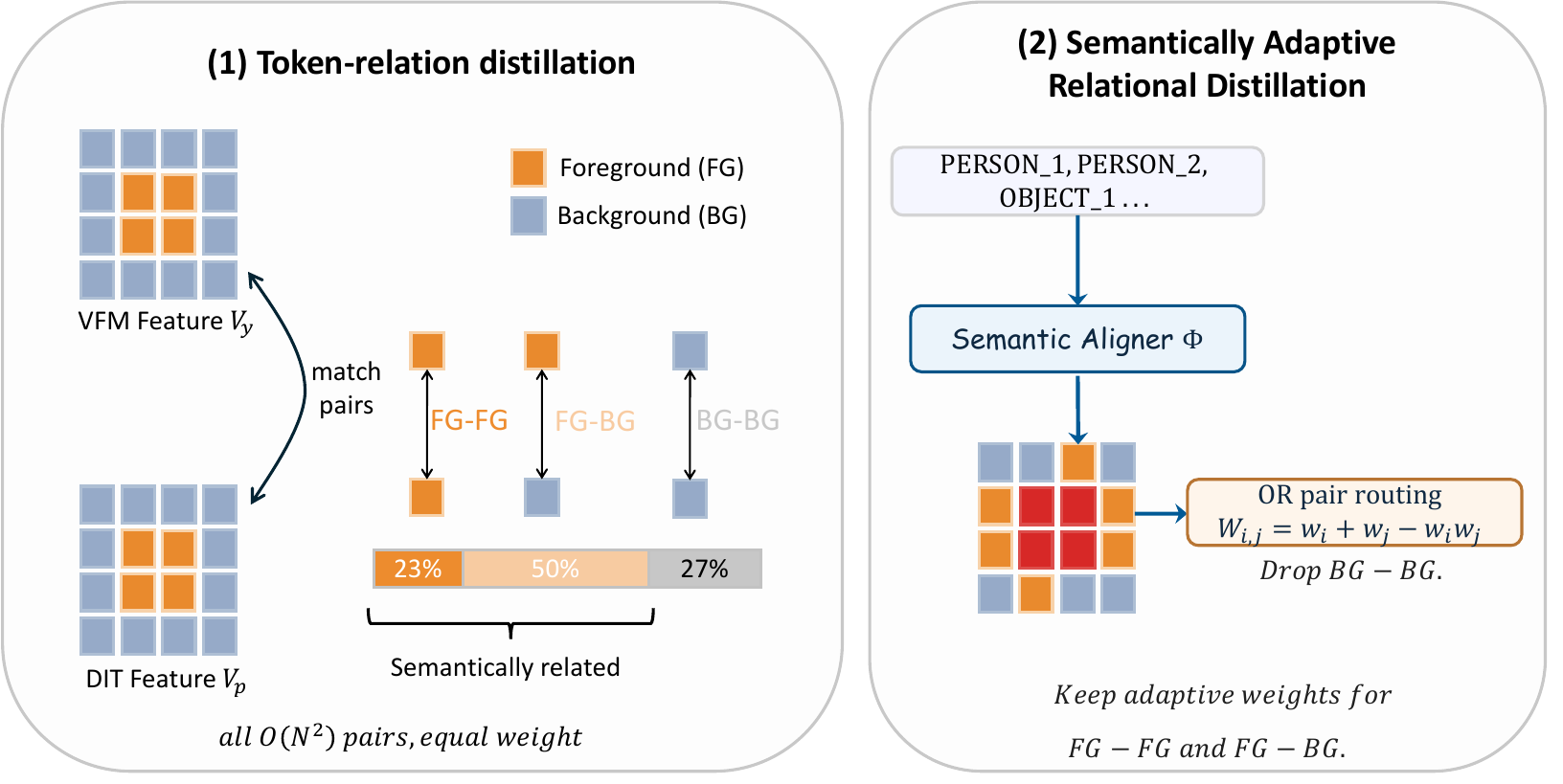}
  \captionsetup{hypcap=false}
  \captionof{figure}{%
    \textbf{SARA makes representation alignment follow the prompt rather than raw pixels.}}
    \label{fig:teaser}
\end{center}

\vspace{2mm}
\begin{center}
{\large\textbf{Abstract}}
\end{center}
Recent video diffusion models (VDMs) synthesize visually convincing clips, yet still drop entities, mis-bind attributes, and weaken the interactions specified in the prompt. Representation-alignment objectives such as VideoREPA and MoAlign improve fine-grained text following by distilling spatio-temporal token relations from a frozen visual foundation model, but their pairwise supervision budget is allocated by visual or motion cues rather than by how relevant each pair is to the prompt. We present \textbf{SARA}, \emph{Semantically Adaptive Relational Alignment}, which keeps token-relation distillation (TRD) on a frozen VFM target and adds a text-conditioned saliency that decides \emph{which} token pairs carry supervision. A lightweight Stage~1 aligner is trained with per-entity SAM~3.1 mask supervision and an InfoNCE regulariser, and its continuous saliency is fused into TRD through a pair-routing operator that assigns each token pair a weight whenever \emph{either} of its two endpoints is salient, thereby routing supervision toward subject-subject and subject-background pairs and away from background-background ones. In the Wan2.2 continual-training setting, SARA improves both text alignment and motion quality over SFT, VideoREPA, and MoAlign on a $13$-dimension VLM rubric, on the public VBench benchmarks, and in a blind user study. Project page: \url{https://saradit.github.io/}.

\section{Introduction}
\label{sec:intro}

Video generation has advanced rapidly in both visual fidelity and temporal coherence. Closed-source systems such as Seedance2~\citep{seedance2026seedance}, Veo3.1~\citep{google2026veo3.1}, and Wan2.7~\citep{wan2.7}, together with open-source models such as LTX2.3~\citep{hacohen2026ltx}, Wan2.2~\citep{wan2025wan}, and HunyuanVideo1.5~\citep{wu2025hunyuanvideo}, can now synthesize videos with realistic appearance and smooth motion. Once visual and motion quality are in place, the remaining bottleneck is faithful prompt following: a generated video is useful to a downstream creator only if it preserves the entities, attributes, interactions, and motion the prompt asks for. Open-source models still fall short here. They miss fine-grained semantic details, bind attributes to the wrong subject, or weaken the interactions specified in the prompt. Fine-grained semantic controllability is therefore a practical requirement when adapting open-source video diffusion models (VDMs).

A natural way to close this gap is continual training on curated video-text data, but the diffusion loss alone is an indirect signal for semantics: it matches pixel-level noise and leaves the DiT to figure out, on its own, which patches correspond to which word in the caption. Representation alignment offers a more direct handle: a frozen visual or video foundation model (VFM) is used as an external reference, and the DiT's hidden states are pulled toward that reference space during training. REPA~\citep{yu2024representation} introduced this for image DiTs, and VideoREPA~\citep{zhang2025videorepa} adapted it to pretrained VDMs by replacing hard per-token alignment with token-relation distillation (TRD), a softer objective that matches pairwise spatial and cross-frame token similarities. Its weakness is one of allocation: VideoREPA weights every token pair equally, so its $O(N^2)$ budget is set by the geometry of the V-JEPA token grid rather than by the caption. But the three kinds of token pairs carry very unequal amounts of semantic content. A background-background (BG-BG) pair relates two patches that no entity in the caption references, so it encodes little prompt-relevant semantics; a subject-background (FG-BG) pair grounds an entity in its surrounding scene, and a subject-subject (FG-FG) pair carries the inter-entity relations a multi-entity prompt is built around, so both are semantically strong. Equal weighting thus spends the budget in inverse proportion to relevance. On our multi-entity training corpus the SAM-derived foreground covers on average only a fraction $p_\text{fg}\!\approx\!0.48$ of the grid, so the budget splits as $(1-p_\text{fg})^2\!\approx\!27\%$ on the semantically weak BG-BG pairs, $2p_\text{fg}(1-p_\text{fg})\!\approx\!50\%$ on FG-BG pairs, and only $p_\text{fg}^2\!\approx\!23\%$ on the FG-FG pairs that most directly encode the prompt (App.~\ref{sec:appendix:pair_budget}). Roughly a quarter of the TRD signal is therefore burnt on pairs the caption never mentions, while the strongest, prompt-defining relations receive the smallest share. MoAlign~\citep{bhowmik2025moalign} addresses this dilution by compressing the VFM target into a flow-supervised motion subspace, which collapses non-moving patches in the target and in effect concentrates supervision on \emph{moving-subject $\leftrightarrow$ moving-subject} pairs. Optical flow is an imperfect saliency proxy, however: it is noisy under occlusion, fast motion, and low-texture regions, it cannot disentangle object motion from camera-induced apparent motion, and it is silent on the static subjects a caption may centrally describe (a person sitting, a cup on a table). The same motion-only restriction also drops every subject-background pair, even though such pairs ground each entity in its scene and carry an independent share of the prompt's semantic content.

The question that motivates SARA is therefore a routing one: \emph{given a fixed $O(N^2)$ pair budget, how should it be allocated so that supervision concentrates on the prompt-relevant relations rather than on background filler?} A caption typically refers to a small subset of the visual content, and the useful semantic signal lives in both subject-subject and subject-background relations~\citep{shi2026vision}, so the answer should route by the prompt rather than by raw pixels or motion. We propose \textbf{SARA}, \emph{Semantically Adaptive Relational Alignment}, a two-stage framework that keeps VideoREPA's VFM target and TRD form unchanged and adds a text-conditioned saliency that tells TRD where to apply its pairwise supervision. Stage~1 trains a lightweight text-conditioned saliency aligner offline from per-entity SAM~3.1 mask supervision~\citep{carion2025sam}, per-entity captions, and an InfoNCE regulariser. Together, the per-entity supervision and InfoNCE prevent the saliency from collapsing onto a fixed foreground prior. Stage~2 freezes the aligner, queries it with the full video caption, and fuses its continuous saliency into token-pair weights through a pair-routing operator (OR by default, so that a pair carries weight whenever either endpoint is salient). This routes TRD away from background-background pairs and toward subject-subject and subject-background relations during continual training of the VDM.

Our contributions are as follows.
\begin{itemize}
\item We recast semantic adaptation for VDMs as a pair-routing problem on top of TRD, formalised through a family of pair-routing operators that decide which token pairs carry supervision (Sec.~\ref{sec:method:stage2}). This view interprets MoAlign as inducing an AND router (a pair is supervised only when both endpoints are salient) through motion presence as an unsupervised saliency proxy, and motivates a text-supervised OR router (either endpoint salient suffices) that also routes supervision to subject-background pairs.
\item We train a lightweight text-conditioned saliency aligner from per-entity SAM~3.1 masks, per-entity captions, and an InfoNCE regulariser, and fuse its continuous output into TRD.
\item Under matched Wan2.2 high-noise continual training, SARA consistently improves over supervised fine-tuning (SFT), VideoREPA, and a MoAlign reproduction on a $13$-dimension vision-language-model (VLM) rubric, on VBench-1.0 and VBench-2.0, and in a blind user study.
\end{itemize}

\section{Related Work}
\label{sec:related}

\paragraph{Video diffusion models.} Text-to-video (T2V) generation has progressed from frame-wise extensions of image diffusion U-Nets~\citep{blattmann2023stable} to large latent diffusion / flow-matching transformers trained on web-scale video-text corpora. Closed-source systems such as Sora~\citep{openai2024sora}, Seedance2~\citep{seedance2026seedance}, Veo3.1~\citep{google2026veo3.1}, Kling3~\citep{kling3}, and Wan2.7~\citep{wan2.7} now produce minutes-long, high-fidelity videos with smooth motion, while open-source counterparts including CogVideoX~\citep{yang2024cogvideox}, LTX2.3~\citep{hacohen2026ltx}, Wan2.2~\citep{wan2025wan}, and HunyuanVideo1.5~\citep{wu2025hunyuanvideo} have closed much of the appearance-quality gap. These open-source models are the dominant base models for downstream continual training. Once architectures and training data scale up, the dominant failure mode shifts from visual fidelity to \emph{fine-grained text following}. Standard benchmarks such as VBench~\citep{huang2024vbench} and the VideoPhy series~\citep{bansal2025videophy} confirm that even SOTA open-source VDMs still drop entities, mis-bind attributes, weaken prompt-specified interactions on multi-subject scenes, and produce physically implausible motion. SARA targets exactly this regime and uses the publicly released Wan2.2 high-noise transformer as the backbone for continual training.

\paragraph{Improving fine-grained semantic alignment in VDMs.} Methods that push a pretrained VDM's prompt fidelity beyond what the base diffusion loss provides split along the standard training stages, and SARA belongs to stage \emph{(ii)} below. \emph{(i) Pre-training / data side.} The pre-training corpus is re-curated and relabelled with VLM rewriters and structured caption formats, so the same diffusion loss carries more semantic signal per gradient step. The open-source VDMs above~\citep{wan2025wan,wu2025hunyuanvideo,hacohen2026ltx} document such data pipelines in their tech reports. \emph{(ii) SFT with auxiliary objectives.} The diffusion loss is kept intact and a representation-alignment term is added that pulls DiT hidden states toward a frozen visual or video foundation encoder (REPA~\citep{yu2024representation}, VideoREPA~\citep{zhang2025videorepa}, MoAlign~\citep{bhowmik2025moalign}, RefAlign~\citep{wang2026refalign}, expanded in the next paragraph). A parallel line instead injects auxiliary modalities such as optical flow, pose, or trajectories during continual training, at the cost of requiring those conditions at inference (e.g.\ Tora~\citep{zhang2025tora}). \emph{(iii) Post-training preference optimization.} Following the RLHF recipe~\citep{ouyang2022training}, the VDM is fine-tuned against a reward model via GRPO-style on-policy exploration that turns the flow-matching ODE~\citep{lipman2022flow} into an SDE~\citep{xue2025dancegrpo}, DPO-style paired classification over preferred / rejected samples~\citep{wallace2024diffusion,liu2025videodpo}, or ReFL-style differentiable-reward back-propagation~\citep{xu2023imagereward}. Post-training is largely orthogonal to SARA's SFT-stage gains, and we leave such combinations to future work.

\paragraph{Representation alignment for diffusion models.} The REPA family is the closest prior art to SARA and shares a single template: regularise a generative DiT by matching a chosen statistic of its hidden states to a frozen visual or video foundation encoder. \emph{REPA}~\citep{yu2024representation} matches each denoiser token to a DINOv2 patch via per-token cosine (refined by REPA-E~\citep{leng2025repae}, which jointly tunes the VAE). \emph{VideoREPA}~\citep{zhang2025videorepa} replaces per-token cosine with TRD on a frozen VideoMAEv2 target (Eqs.~\eqref{eq:trd-sim}--\eqref{eq:trd-base}). \emph{MoAlign}~\citep{bhowmik2025moalign} keeps TRD but compresses $V_y$ into a flow-supervised motion subspace $\Phi_\text{mot}$ and decays the cross-frame term by $\exp(-|t-u|/\tau)$, biasing supervision toward moving patches. \emph{RefAlign}~\citep{wang2026refalign} adapts the template to the reference-to-video setting with a contrastive DINOv3 loss between reference-branch tokens and the target. These methods vary how the alignment is shaped (per-token vs.\ relational, appearance vs.\ motion, image- vs.\ text-conditioned), but none lets the text prompt decide \emph{which} pairs carry supervision.

SARA adds an orthogonal ingredient: the routing of the alignment loss itself. It reuses VideoREPA's TRD form on a frozen VFM target and shifts the shaping signal to a text-supervised saliency trained with per-entity SAM~3.1 masks and an InfoNCE regulariser (Sec.~\ref{sec:method:stage1}). Within this view, VideoREPA is the constant-saliency limit, while MoAlign can be interpreted as an AND pair-routing operator that biases supervision toward moving subject-subject pairs. SARA's default OR pair-routing operator additionally keeps subject-background pairs and consistently improves over both alternatives~(Sec.~\ref{sec:exp}).

\section{Method}
\label{sec:method}

\begin{figure}[htbp]
\centering
\includegraphics[width=\linewidth]{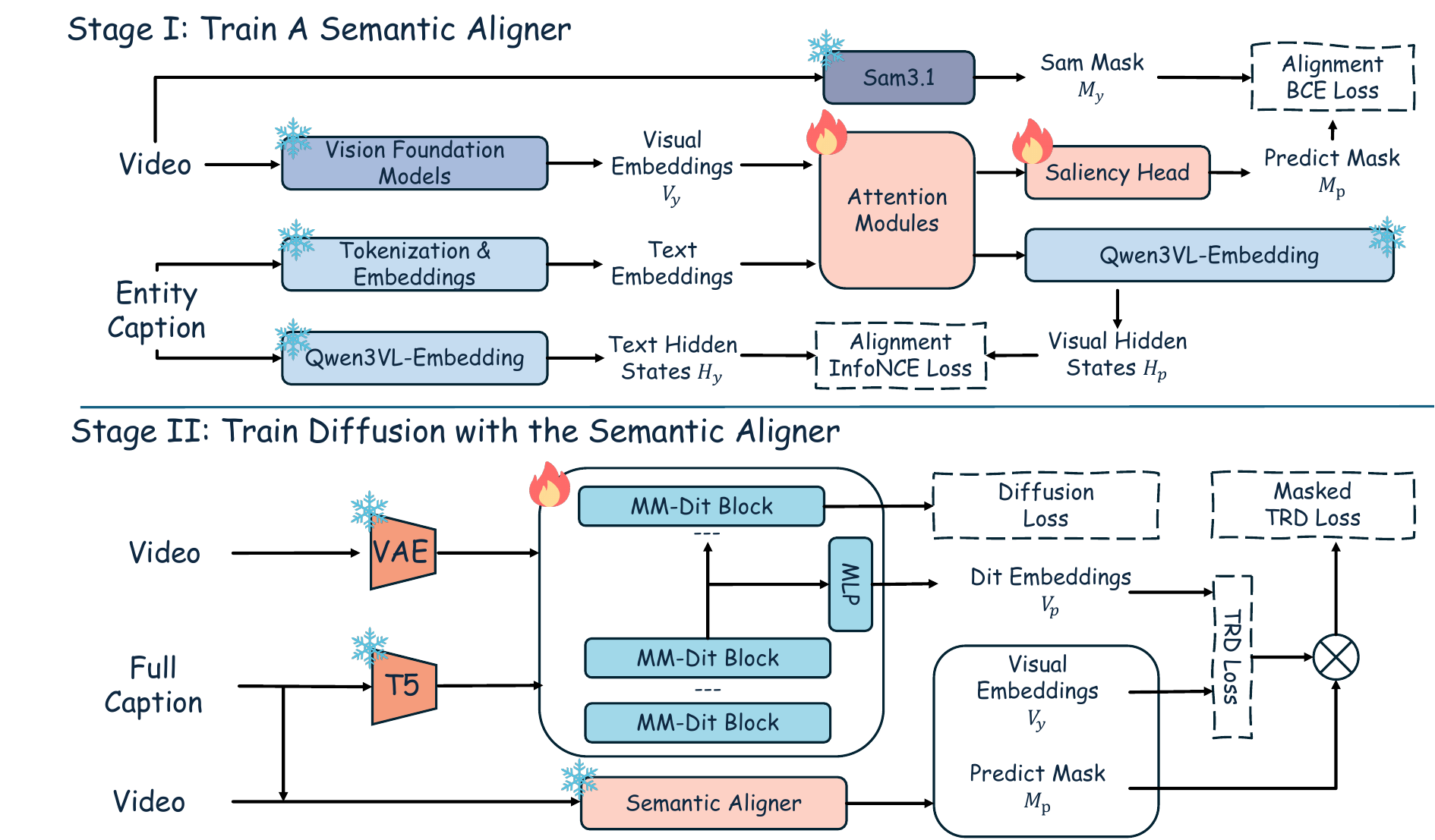}
\caption{\textbf{Overview of SARA.} \emph{Stage~I (top):} a lightweight aligner on top of frozen V-JEPA, SAM~3.1, and Qwen3-VL-Embedding backbones learns, for any (video, caption) pair, a text-conditioned per-patch saliency $M_p$, supervised jointly by per-entity, combined-entity, and background SAM masks ($\mathcal{L}_\text{BCE}$) and calibrated by a caption-level InfoNCE. \emph{Stage~II (bottom):} the frozen aligner is queried with the full caption, and its saliency is turned into pair weights that route a masked token-relation distillation loss, added to the diffusion loss of a trainable DiT.}
\label{fig:overview}
\end{figure}

SARA decouples \emph{where} relational alignment should be applied from \emph{how} it is computed. We first recall the TRD formulation underlying SARA and define the entity vocabulary used throughout the paper (Sec.~\ref{sec:method:prelim}). We then identify the routing gap in vanilla TRD and introduce two design choices to address it (Sec.~\ref{sec:method:motivation}). Finally, we train a lightweight text-conditioned saliency aligner to realise these choices (Sec.~\ref{sec:method:stage1}) and freeze it to route TRD on the Wan2.2 high-noise VDM (Sec.~\ref{sec:method:stage2}). An overview of our pipeline is shown in Fig.~\ref{fig:overview}.

\subsection{Preliminaries}
\label{sec:method:prelim}

\paragraph{Latent video diffusion.} A latent VDM~\citep{yang2024cogvideox,wan2025wan} generates videos in the latent space of a frozen 3D VAE. Given a clean video $x_0 \in \mathbb{R}^{F\times H\times W\times C}$ and a text condition $c$, the VAE produces a latent $z_0$, and a denoising transformer $\epsilon_\theta$ is trained under the standard flow-matching / diffusion objective $\mathcal{L}_\text{diff}(\theta) = \mathbb{E}_{t,z_0,\epsilon} \bigl[\|\epsilon-\epsilon_\theta(z_t,t,c)\|_2^2\bigr]$, with $z_t$ a noisy version of $z_0$ at timestep $t$.

\paragraph{Token-relation distillation.} REPA~\citep{yu2024representation} aligns each denoiser token with a frozen visual encoder feature via per-token cosine similarity. As argued by VideoREPA~\citep{zhang2025videorepa}, this hard alignment is unsuitable for fine-tuning pretrained VDMs and ignores temporal dynamics. VideoREPA instead matches pairwise token similarities between a projected DiT hidden state $V_p \in \mathbb{R}^{B\times T\times N\times D}$ and the VFM features $V_y \in \mathbb{R}^{B\times T\times N\times D}$ (interpolated to a common $T, N = hw$ grid, with $t,u \in [T]$ indexing frames and $i,j \in [N]$ indexing spatial token positions). With $\hat V_p$, $\hat V_y$ denoting L2-normalized features, the within-frame (spatial) and cross-frame (temporal) cosine similarities are
\begin{equation}
S^{X}_{t,i,j}=\hat V_{X,t,i} \hat V_{X,t,j}^\top,\qquad C^{X}_{t,i,u,j}=\hat V_{X,t,i} \hat V_{X,u,j}^\top,\qquad X\in\{p,y\},
\label{eq:trd-sim}
\end{equation}
giving stacked spatial Gram matrices $S^X \in \mathbb{R}^{T\times N\times N}$ and a cross-frame Gram tensor $C^X \in \mathbb{R}^{T\times N\times T\times N}$. The TRD loss sums the within-frame and cross-frame L1 differences~\citep{zhang2025videorepa}:
\begin{equation}
\mathcal{L}_\text{TRD} = \underbrace{\tfrac{1}{T N^2} \sum_{t,i,j} \bigl|S^{y}_{t,i,j}-S^{p}_{t,i,j}\bigr|}_{\text{Spatial component}}+\underbrace{\tfrac{1}{T(T-1)N^2}  \sum_{\substack{t\neq u\\i,j}} \bigl|C^{y}_{t,i,u,j}-C^{p}_{t,i,u,j}\bigr|}_{\text{Temporal component}}.
\label{eq:trd-base}
\end{equation}
MoAlign~\citep{bhowmik2025moalign} extends TRD by attaching an exponential temporal-distance decay $\omega_{t,u}=\exp(-|t-u|/\tau)$ to the cross-frame term and swapping $V_y$ for a flow-supervised motion subspace.

\paragraph{MTSS entities.} We caption every video in the Multi-Stream Scene Script (MTSS) format of~\citet{team2026script}, which factorises a clip into per-entity descriptions linked by stable identifiers and is therefore a natural source of per-entity supervision (App.~\ref{sec:appendix:mtss} gives the construction). From each MTSS caption SARA extracts (i)~$K$ \emph{entity captions} $c_k$, each paired with a binary \emph{entity mask} $M_k$ obtained offline from a frozen segmentation backbone $E_s$ (instantiated in Sec.~\ref{sec:method:stage1}, pipeline in App.~\ref{sec:appendix:simplifier}), (ii)~the foreground concatenation $c_\text{fg} = [c_1;\dots;c_K]$ with union mask $M_\text{fg}{=}\bigcup_k M_k$, (iii)~a background caption $c_\text{bg}$ with complement mask $M_\text{bg} = \mathbf{1} - M_\text{fg}$, and (iv)~the \emph{full caption} $c$ that serialises all streams. Stage~1 trains the aligner on (i)--(iii) only. The full caption $c$ is used solely at Stage~2 inference, where both the VDM and the frozen aligner are conditioned on it.

\subsection{Motivation and design}
\label{sec:method:motivation}

VideoREPA's TRD in Eq.~\eqref{eq:trd-base} weights every token pair equally, so its $O(N^2)$ budget is dominated by background-background pairs and the supervision on the few prompt-relevant pairs is diluted (Fig.~\ref{fig:pair_budget}). MoAlign re-allocates by projecting the VFM target into a flow-supervised motion subspace, which suppresses non-moving patches and biases supervision toward moving-subject pairs, an AND routing effect under a motion-presence proxy. Optical flow is itself a noisy estimator that mis-handles occlusion, fast motion, low-texture regions, and camera-induced apparent motion. Even when accurate, it is silent on the static subjects a caption may centrally describe, and the same restriction under-emphasises subject-background pairs, a major source of fine-grained semantic grounding. SARA replaces this implicit, motion-only bias with an explicit, text-supervised saliency, while keeping V-JEPA as the TRD target.

The mechanism is a text-conditioned saliency, predicted from V-JEPA tokens fused with the caption through cross-attention and shaped by two complementary auxiliary losses. A \emph{local, mask-anchored} BCE focuses the head on caption-mentioned patches by grounding the fused features in the per-patch entity masks $M_k$ from $E_s$, following LaST-ViT~\citep{shi2026vision} in placing the useful semantic signal in foreground-background relations. A \emph{global, embedding-space} InfoNCE prevents collapse onto a dominant subject and preserves cross-entity contrast by aligning the fused features back to the caption hidden state, following VL-JEPA~\citep{chen2025vl}. Without either ingredient, the predicted saliency degenerates: ablating the InfoNCE regulariser concentrates $M_p$ on the dominant subject, and replacing the entity-separated $K + 2$ supervision by a single union-mask forward saturates $M_p$ across the entire foreground (Fig.~\ref{fig:stage1_ablation}, App.~\ref{sec:appendix:viz_ablation}). Tab.~\ref{tab:stage1_diagnostics} quantifies both collapses.

\subsection{Stage 1: Text-conditioned saliency aligner}
\label{sec:method:stage1}

\paragraph{Frozen backbones.} Stage~1 uses three frozen backbones: a video encoder $E_v$ (V-JEPA~2.1~\citep{mur2026v}) producing visual embeddings $V_y=E_v(x_0)\in\mathbb{R}^{B\times N_v\times D_v}$, a segmentation agent $E_s$ (SAM~3.1 Multiplex~\citep{carion2025sam}) that returns one binary mask $M_k$ per detected entity from a noun prompt, and a text model $E_t$ (Qwen3-VL-Embedding~\citep{qwen3vlembedding}), used in two disjoint modes. Its input-embedding lookup $E_\text{emb}$ produces per-token embeddings $\tilde E=E_\text{emb}(c)$ consumed by cross-attention, and its full transformer stack $E_\text{lm}$ produces contextualised hidden states consumed by InfoNCE. All three remain frozen throughout Stage~1 and Stage~2. Model variants and shapes are in App.~\ref{sec:appendix:training}.

\paragraph{Aligner architecture.} Three trainable modules sit on top of the frozen backbones. A stack $\Phi_\text{CA}$ of cross- and self-attention blocks fuses $V_y$ with the caption. Cross-attention takes visual queries and text keys/values $\tilde E$, and outputs text-enhanced features
\begin{equation}
V'_y = \Phi_\text{CA}(V_y, \tilde E) \in \mathbb{R}^{B\times N_v\times D_v}.
\end{equation}
A saliency head $\Phi_\text{sal}$ (MLP + sigmoid) produces a per-patch saliency mask
\begin{equation}
M_p = \sigma\bigl(\Phi_\text{sal}(V'_y)\bigr) \in [0,1]^{B\times N_v},
\label{eq:saliency}
\end{equation}
and a visual projector $\Phi_\text{proj}$ maps $V'_y$ into the input-embedding space of $E_\text{lm}$ so that $E_\text{lm}$ can consume it in inputs-embeds mode for the InfoNCE objective below. Block counts, MLP hidden sizes, and normalisation choices are in App.~\ref{sec:appendix:training}.

\begin{figure}[t]
\centering
\begin{subfigure}[t]{0.49\linewidth}\centering
  \includegraphics[width=\linewidth]{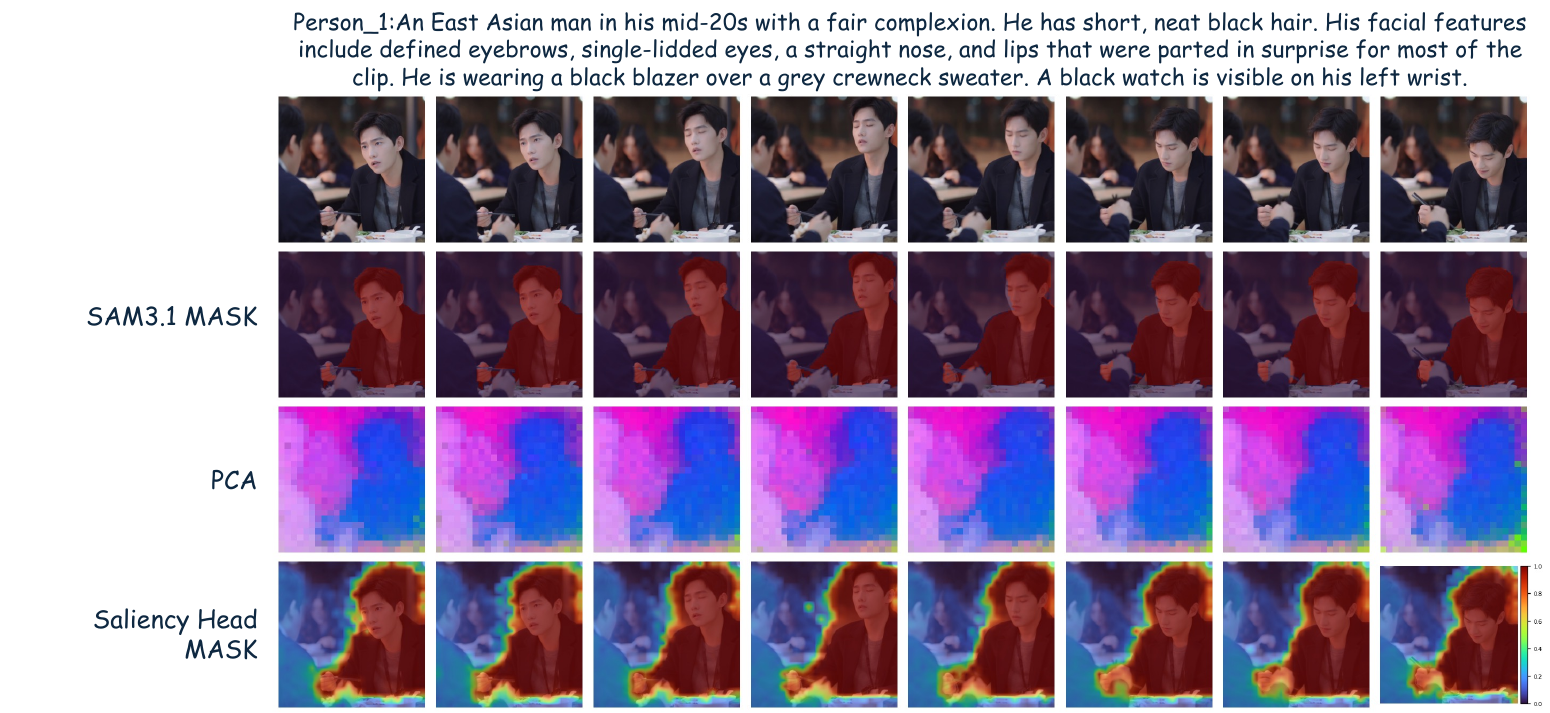}
  \caption{Query $c_1$: \texttt{PERSON\_1}.}
  \label{fig:viz_stage1:p1}
\end{subfigure}\hfill
\begin{subfigure}[t]{0.49\linewidth}\centering
  \includegraphics[width=\linewidth]{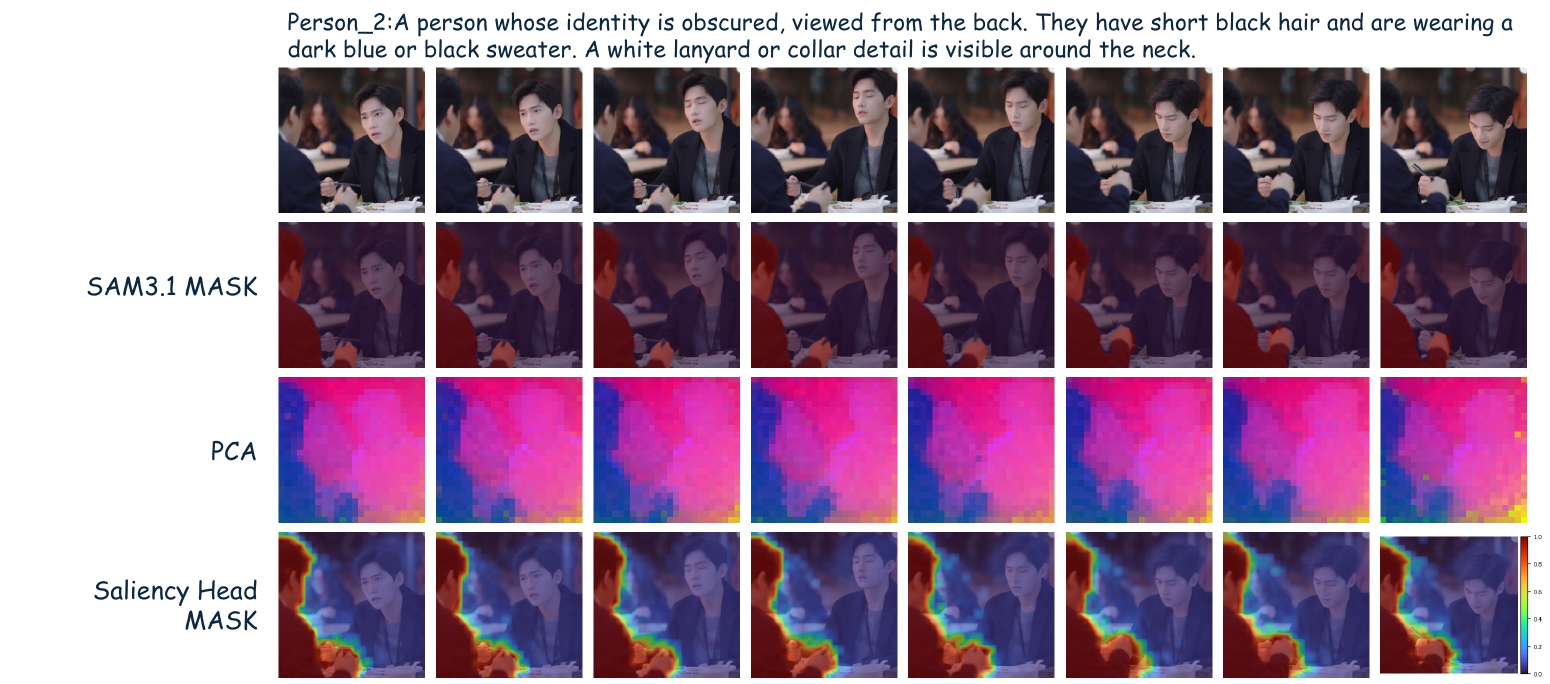}
  \caption{Query $c_2$: \texttt{PERSON\_2}.}
  \label{fig:viz_stage1:p2}
\end{subfigure}\\[0.4em]
\begin{subfigure}[t]{0.49\linewidth}\centering
  \includegraphics[width=\linewidth]{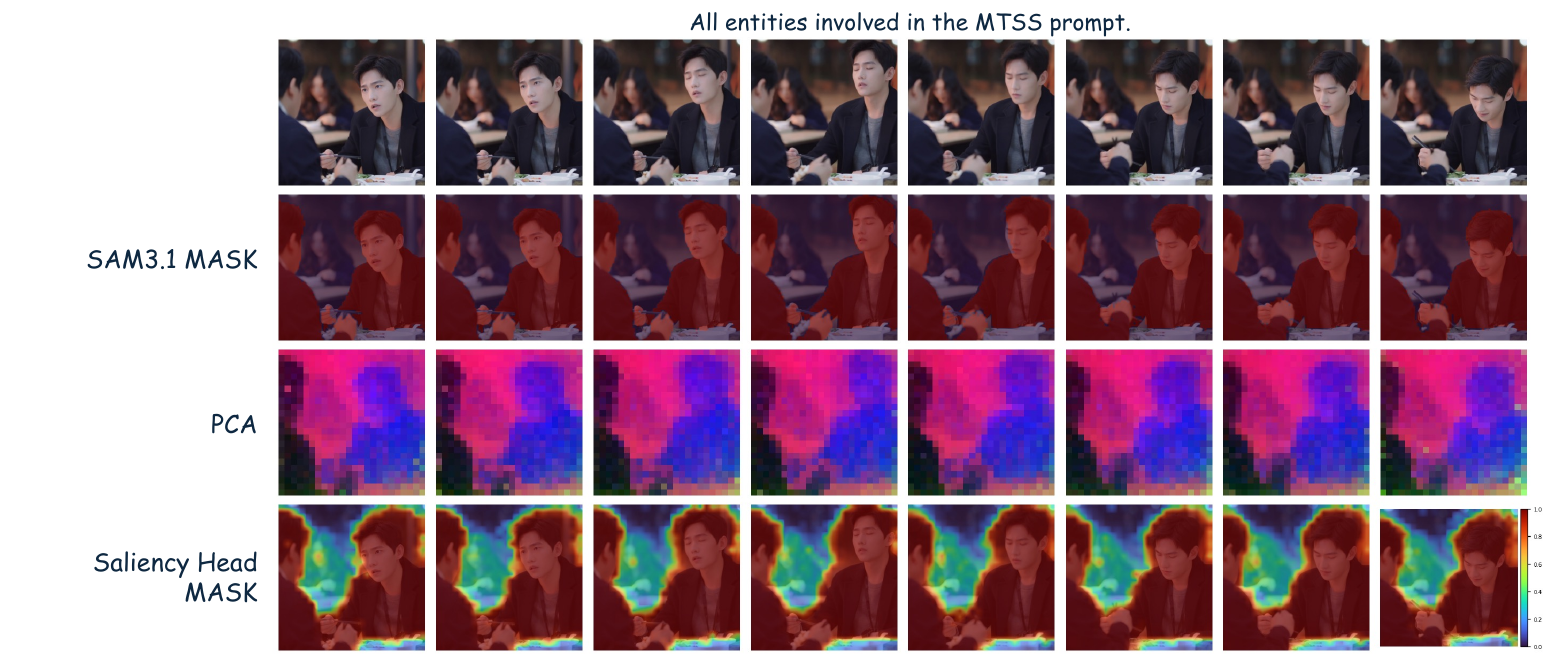}
  \caption{Query $c_\text{fg} = [c_1;c_2;\dots]$: combined-entity.}
  \label{fig:viz_stage1:all}
\end{subfigure}\hfill
\begin{subfigure}[t]{0.49\linewidth}\centering
  \includegraphics[width=\linewidth]{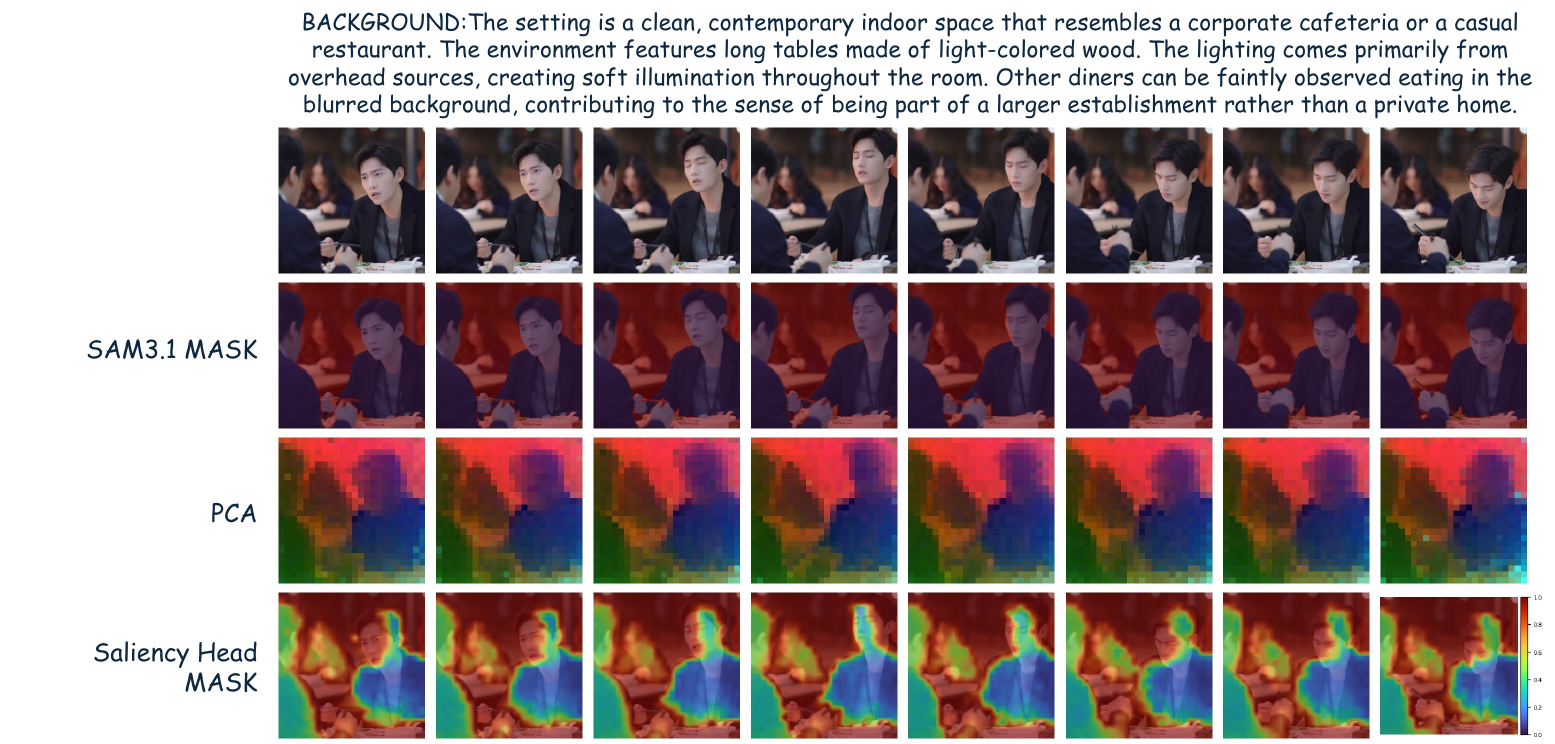}
  \caption{Query $c_\text{bg}$: background (cafeteria setting).}
  \label{fig:viz_stage1:bg}
\end{subfigure}
\caption{\textbf{Stage~1 saliency on the four supervision-time query types}, eight frames of one held-out clip. Rows in each panel: input frames, SAM~3.1 reference mask $M_y$, PCA of $V'_y$, predicted saliency $M_p$ (Eq.~\eqref{eq:saliency}; jet colormap, redder = higher). Under the two per-entity queries the head selects different tokens for the two co-located persons (panels a--b), softly unions them under the combined-entity query (c), and inverts onto the scene under the background query (d), so the routing is genuinely text-conditioned rather than a fixed foreground prior.}
\label{fig:viz_stage1}
\end{figure}

\paragraph{Mask-anchored BCE with entity-separated supervision.} We bilinearly downsample any target binary mask $M$ to the V-JEPA spatial grid $H_s \times W_s$ and broadcast across the $T_s$ temporal positions to obtain a per-patch target $M_y\in\{0,1\}^{N_v}$ aligned with $V_y$. The mask loss is per-patch binary cross-entropy~(BCE) between the saliency prediction and this target,
\begin{equation}
\mathcal{L}_\text{BCE} = \mathrm{BCE}\bigl(M_p,\,M_y\bigr).
\label{eq:lbce}
\end{equation}
The choice of $M$ matters: conditioning the aligner on a single global caption with the union mask collapses $\Phi_\text{sal}$ onto a fixed foreground prior, since both query and target then stay constant across all Reference items of a video. SARA instead instantiates $M$ and the conditioning caption at three granularities, sharing parameters across $K + 2$ forwards per video: (i) \emph{$K$ per-entity} forwards using $(c_k, M_k)$; (ii) one \emph{combined-entity} forward using $(c_\text{fg}, M_\text{fg})$ with $c_\text{fg} =[c_1;\dots;c_K]$ and $M_\text{fg} =\bigcup_k M_k$; and (iii) one \emph{background} forward using the SCENE-stream caption $c_\text{bg}$ and complement mask $M_\text{bg} =\mathbf{1} -M_\text{fg}$. The per-entity forwards prevent the foreground-prior collapse, while the combined-entity and background forwards anchor the saliency at the foreground-background level (Fig.~\ref{fig:sam3_entity_masks}, App.~\ref{sec:appendix:viz_sam3}). Sweeping these four supervision-time queries on a held-out clip (Fig.~\ref{fig:viz_stage1}) confirms the intended behaviour: the trained aligner places $M_p$ on different V-JEPA tokens for the two persons of the same scene under $c_1$ vs.\ $c_2$, covers both as a soft union under $c_\text{fg}$, and cleanly inverts under the background query $c_\text{bg}$; its PCA row further shows $\Phi_\text{CA}$ already organises features into entity-specific subspaces that the saliency head reads off rather than re-discovers. The full MTSS caption that aggregates Shot, Event, and Global streams is never seen at this stage; Sec.~\ref{sec:method:stage2} (Fig.~\ref{fig:viz_stage2}) shows the aligner generalises compositionally to it at Stage~2 inference.

\paragraph{Embedding-space InfoNCE loss.} Unlike the cross-attention side, which only consumes $\tilde E$, this InfoNCE loss operates on the \emph{hidden states} of the full language model $E_\text{lm}$. The projected features $\Phi_\text{proj}(V'_y)$ live in $E_\text{lm}$'s input-embedding space, so we push them through $E_\text{lm}$ in inputs-embeds mode and last-token-pool its output to give an $L_2$-normalised visual hidden state $H_p$. The caption used in this forward (per-entity $c_k$, combined $c_\text{fg}$, or background $c_\text{bg}$) is tokenized and runs through the same frozen $E_\text{lm}$, and last-token-pooled to give the text hidden state $H_y$. With temperature $\tau_\text{nce}$ and batch size $B$,
\begin{equation}
\mathcal{L}_\text{InfoNCE} = -\frac{1}{B}\sum_{i=1}^{B}\log\frac{\exp(H_{p,i}^\top H_{y,i} / \tau_\text{nce})}{\sum_{j=1}^{B}\exp(H_{p,i}^\top H_{y,j} / \tau_\text{nce})}.
\label{eq:lnce}
\end{equation}
Both indices run over the $B$ (video, caption) forwards in the mini-batch: $i$ selects the \emph{anchor} forward, whose visual hidden state $H_{p,i}$ is contrasted against the caption hidden states $H_{y,j}$ of every forward $j$ in the same batch. The single positive is the diagonal term $j=i$ (the caption that actually conditioned forward $i$), and the $B-1$ off-diagonal terms $j\neq i$ act as in-batch negatives. Because a video contributes one forward per caption granularity ($c_k$, $c_\text{fg}$, $c_\text{bg}$), these negatives include the \emph{other entities and the background of the same clip}, so minimising Eq.~\eqref{eq:lnce} drives each visual state toward its own caption while keeping different entities of one scene mutually contrastive, preventing the saliency from collapsing onto a single dominant subject.

\paragraph{Stage 1 objective.} The aligner is trained with
\begin{equation}
\mathcal{L}_\text{stage1} = \lambda_\text{BCE}\,\mathcal{L}_\text{BCE} + \lambda_\text{InfoNCE}\,\mathcal{L}_\text{InfoNCE},\qquad \lambda_\text{BCE}=\lambda_\text{InfoNCE}=1.
\label{eq:lstage1}
\end{equation}
Only $\Phi_\text{CA}$, $\Phi_\text{sal}$, and $\Phi_\text{proj}$ receive gradients, while $E_v$, $E_s$, $E_\text{emb}$, and $E_\text{lm}$ remain frozen. Fig.~\ref{fig:stage1_ablation} (App.~\ref{sec:appendix:viz_ablation}) and the matched quantitative metrics in Tab.~\ref{tab:stage1_diagnostics} show that each ingredient of Eq.~\eqref{eq:lstage1} is necessary at the Stage~1 level, and Sec.~\ref{sec:exp:ablation} confirms this on the downstream VLM rubric.

\subsection{Stage 2: Saliency-routed TRD}
\label{sec:method:stage2}

\begin{figure}[t]
\centering
\includegraphics[width=0.82\linewidth]{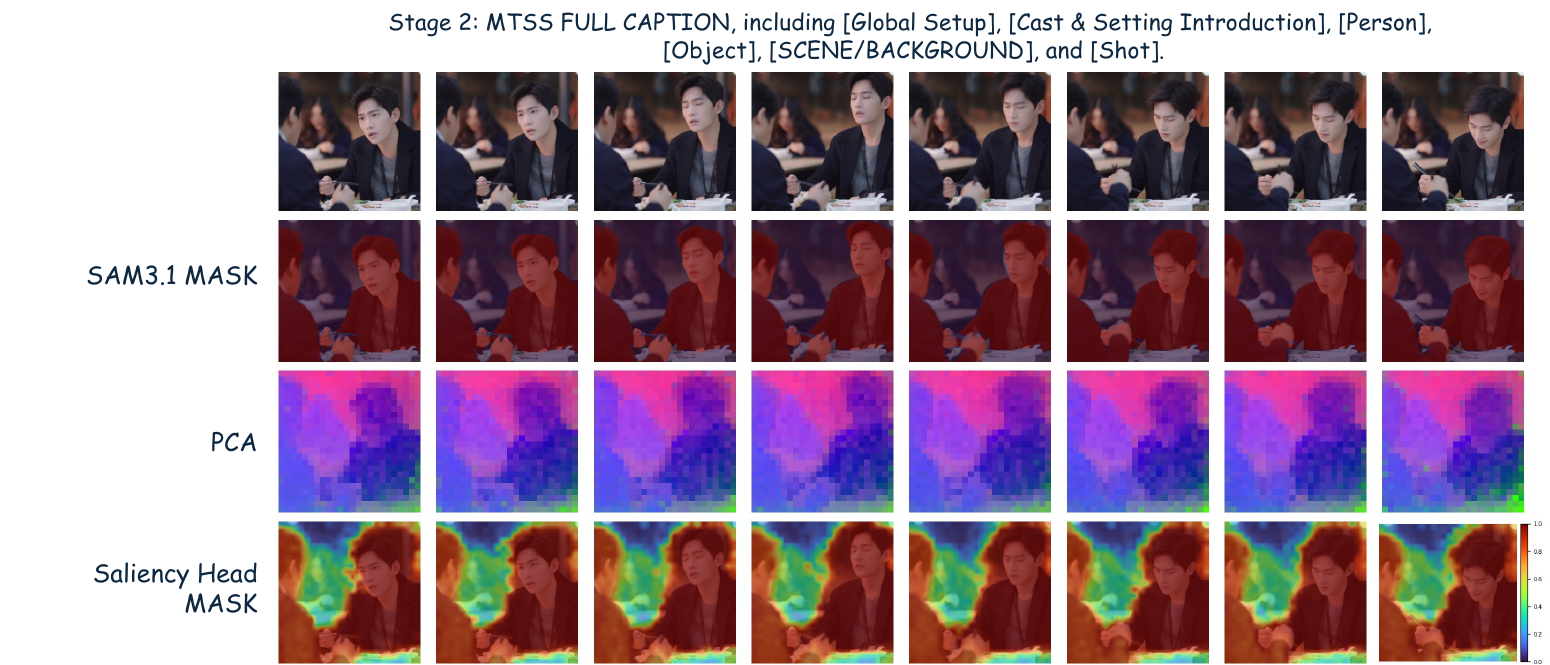}
\caption{\textbf{Stage~2 saliency on the full MTSS caption.} Same aligner and clip as Fig.~\ref{fig:viz_stage1}, queried with the full MTSS string $c$ that the VDM also consumes at TRD time. Rows: input frames, SAM~3.1 union mask $M_\text{fg}$ (reference only, not used at Stage~2), PCA of $V'_y$, predicted saliency $M_p$ (jet colormap, redder = higher). Although never trained on the concatenated MTSS string, the aligner covers both named subjects in one response that tracks $M_\text{fg}$, and it stays graded: highest on the subjects, intermediate on nearby background (the table, the wall behind), lowest on far-field background, exactly what the OR weight $W^{\vee}$ of Eq.~\eqref{eq:pair-ops} needs to grade rather than gate each subject-background pair.}
\label{fig:viz_stage2}
\end{figure}

\paragraph{Inference-time saliency.} The frozen aligner in Stage~1 is fed the same full video caption $c$ as the VDM and emits a continuous saliency $M_p(x_0,c)\in[0,1]^{B\times T_s\times H_s W_s}$ on the V-JEPA grid. Although trained only on Reference-stream captions, the aligner generalises compositionally to the full MTSS string and attends to all named entities jointly (Fig.~\ref{fig:viz_stage2}): the response under $c$ is super-additive over the per-entity responses of Fig.~\ref{fig:viz_stage1}, closely tracks the SAM-derived foreground union, and adaptively grades the background, with intermediate values on tokens spatially or semantically close to a named subject. This grading, rather than a hard binary mask, is what the OR weight $W^{\vee}$ below needs to keep every subject-background pair while ranking it by background relevance. Since the TRD target reuses the same $E_v$, $M_p$ indexes exactly the patches TRD aligns, and the pair-routing operator below turns this continuous grading into per-pair routing strength.

\paragraph{Pair-weight construction.} We define a \emph{pair-routing operator} as any function that maps the per-token saliency $w = M_p \in [0,1]^{B\times T_s\times N}$ to a per-pair weight $W_{ij}\!\in\![0,1]$ that decides how much TRD supervision the pair $(i,j)$ receives. We instantiate three pair-routing operators as fuzzy-logic relaxations of the corresponding Boolean operations on the binary saliency:
\begin{equation}
\underbrace{W^{\wedge}_{ij} = w_i\,w_j}_{\text{AND}},\qquad
\underbrace{W^{\vee}_{ij} = w_i + w_j - w_i\,w_j}_{\text{OR}},\qquad
\underbrace{W^{\oplus}_{ij} = |w_i - w_j|}_{\text{XOR}}.
\label{eq:pair-ops}
\end{equation}
Let FG and BG denote foreground and background tokens, respectively. In the discrete limit where $w \in \{0,1\}$, the relation space is cleanly separated: $W^{\wedge}$ only retains FG-FG pairs, $W^{\vee}$ additionally includes FG-FG and FG-BG pairs, and $W^{\oplus}$ only keeps FG-BG boundary pairs. Keeping $w$ continuous preserves the InfoNCE calibration of Eq.~\eqref{eq:lnce} and yields differentiable gradients via the saliency-weighted denominators in Eqs.~\eqref{eq:lspa}--\eqref{eq:ltmp}. The constant-saliency limit $w \equiv 1$ gives vanilla VideoREPA, while MoAlign's flow-supervised motion subspace can be viewed as a separate mechanism that induces an AND bias toward moving FG-FG relations. SARA's default $W^{\vee}$ instead uses text-supervised saliency to cover the FG-BG pairs that AND drops.

\paragraph{Masked TRD loss.} Let $S^X, C^X$ be the similarities of Eq.~\eqref{eq:trd-sim} with $V_p$ the projected DiT embeddings and $V_y=E_v(x_0)$ the Stage~1 VFM embeddings. With temporal decay $\omega_{t,u}=\exp(-|t-u|/\tau)$, the masked TRD loss replaces the uniform averages of Eq.~\eqref{eq:trd-base} by saliency-weighted ones:
\begin{align}
\mathcal{L}_\text{m\text{-}TRD}^{\text{spa}} &= \frac{\sum_{t,i,j}\,W^{\vee}_{t,i,j}\,\bigl|S^{y}_{t,i,j}-S^{p}_{t,i,j}\bigr|}{\sum_{t,i,j}\,W^{\vee}_{t,i,j}+\varepsilon}, \label{eq:lspa}\\
\mathcal{L}_\text{m\text{-}TRD}^{\text{tmp}} &= \frac{\sum_{t\neq u,i,j}\,\omega_{t,u}\,W^{\vee}_{t,i,u,j}\,\bigl|C^{y}_{t,i,u,j}-C^{p}_{t,i,u,j}\bigr|}{\sum_{t\neq u,i,j}\,\omega_{t,u}\,W^{\vee}_{t,i,u,j}+\varepsilon}, \label{eq:ltmp}\\
\mathcal{L}_\text{m\text{-}TRD} &= \mathcal{L}_\text{m\text{-}TRD}^{\text{spa}} + \lambda_\text{tmp}\,\mathcal{L}_\text{m\text{-}TRD}^{\text{tmp}}, \label{eq:lmtrd}
\end{align}
where the OR pair weight is the fuzzy-OR of the two endpoint saliencies of each token pair,
\begin{equation}
W^{\vee}_{t,i,u,j}=w_{t,i}+w_{u,j}-w_{t,i}\,w_{u,j},\qquad W^{\vee}_{t,i,j} \equiv W^{\vee}_{t,i,t,j}=w_{t,i}+w_{t,j}-w_{t,i}\,w_{t,j},
\label{eq:Wor}
\end{equation}
the second form is the within-frame special case $u = t$ used in Eq.~\eqref{eq:lspa}. Setting $w \equiv 1$ recovers VideoREPA's TRD up to normalisation, and adding a finite $\tau$ isolates the temporal-decay component used by MoAlign without matching its motion-subspace target. We default to $\tau = \infty$, and confirm in Sec.~\ref{sec:exp:ablation} that finite $\tau$ is not the source of SARA's gains.

\paragraph{Stage 2 objective.} Wan2.2 is continually trained with
\begin{equation}
\mathcal{L}_\text{stage2} = \mathcal{L}_\text{diff} + \lambda_\text{TRD}\,\mathcal{L}_\text{m\text{-}TRD}.
\label{eq:lstage2}
\end{equation}
Only the VDM and the small TRD projector receive gradients, while $E_v$, $E_s$, $E_\text{emb}$, $E_\text{lm}$, and the entire Stage~1 aligner ($\Phi_\text{CA}, \Phi_\text{sal}, \Phi_\text{proj}$) remain frozen, so SARA adds no trainable parameters to the diffusion path beyond the standard REPA projector.

\section{Experiments}
\label{sec:exp}

\subsection{Setup}
\label{sec:exp:setup}

\paragraph{Dataset.} We start from an internal pool of $\sim 4$M multi-subject video clips. Every clip is first recaptioned into MTSS form by the pipeline of App.~\ref{sec:appendix:mtss} and then ranked by its \emph{entity count}, defined as the number of Reference items whose type is \texttt{PERSON\_*} or \texttt{OBJECT\_*} (\texttt{SCENE\_*} items are excluded from the count). The $500$K clips with the highest entity count form the training corpus used throughout the paper, and a fixed $800$-clip test set is held out from the top of the same ranking (mean entity count $\sim 5.2$ \texttt{PERSON}/\texttt{OBJECT} per clip, versus $\sim 1.8$ on a uniformly sampled subset). Training and evaluation are therefore concentrated in the multi-entity regime that SARA's saliency routing targets. The same corpus supplies both Stage~1 saliency-aligner training and Stage~2 VDM continual training, so all methods see identical data. Ground-truth entity masks for Stage~1 are produced offline by the frozen SAM~3.1 Multiplex using the per-Reference \texttt{semantic\_description}s simplified as in App.~\ref{sec:appendix:simplifier}.

\paragraph{Backbone and training.} All baselines and SARA are continually trained on the same Wan2.2 high-noise VDM under an identical schedule, including the optimiser, batch size, GPU count, and number of steps. Only the auxiliary objective changes. Detailed hyperparameters are listed in Tables~\ref{tab:appendix:hparams-stage1} and~\ref{tab:appendix:hparams-stage2} (App.~\ref{sec:appendix:training}).

\paragraph{Baselines.} We compare four approaches: (i) the \textbf{pretrained Wan2.2} high-noise model without continual training; (ii) \textbf{SFT} with only the diffusion loss; (iii) \textbf{VideoREPA}~\citep{zhang2025videorepa}, TRD on V-JEPA~2.1 features without saliency routing; and (iv) a \textbf{MoAlign}~\citep{bhowmik2025moalign} reproduction with motion subspace $D_m=64$ and temporal decay $\tau=10$. To isolate the routing mechanism from the VFM target, in our reproduced VideoREPA we replace the original VideoMAEv2 backbone with V-JEPA~2.1. As a result, all three REPA-family variants (VideoREPA, MoAlign, SARA) share a frozen V-JEPA~2.1 target (MoAlign additionally projects $V_y$ through its motion subspace). The original source videos are reported as an oracle upper bound. All methods use fixed-resolution training.

\paragraph{VLM-rubric protocol.} The $800$-clip test set is scored by three independent VLM judges (Qwen3.5-27B~\citep{qwen2026qwen35}, Qwen3.6-35B-A3B~\citep{qwen2026qwen36}, Gemma-4-31B-it~\citep{google2026gemma4}) on $13$ rubric dimensions (six text-alignment TA, seven motion-quality MQ, $1$--$5$ each). Using three judges rather than one reduces per-grader bias. We aggregate across judges in two ways: the per-dimension average (\emph{mean}) and the per-dimension majority vote (\emph{vote}, ties broken upward), and the TA / MQ columns average the respective sub-dimensions. We cross-check the rubric against two independent protocols, public VBench-1.0 / 2.0 (Sec.~\ref{sec:exp:vbench}) and a blind pairwise user study (Sec.~\ref{sec:exp:userstudy}), and the rankings agree across all three. Full rubric text, judge configuration, and per-sub-dimension scores are in App.~\ref{sec:appendix:vlmrubric}.

\subsection{Main comparison}
\label{sec:exp:main}

Table~\ref{tab:main} reports the VLM-rubric scores. SFT improves text alignment but slightly degrades motion, while VideoREPA and MoAlign recover part of the motion gap but still trail SARA on both dimensions. Among the four continually-trained methods, SARA is the only one that improves both dimensions at once under \emph{mean} and \emph{vote}, and it beats the strongest matched-setting baseline (MoAlign) on every column. The \emph{real video} row is a protocol-level ceiling, not a $5.0$ saturation point, and caption-rewriter and judge-VLM noise affect every row equally (App.~\ref{sec:appendix:vlmrubric}). Relative to that ceiling, SARA closes more of the gap to the strongest baseline than any other row.

\begin{table}[htbp]
\centering
\caption{\textbf{VLM-rubric main comparison} on Wan2.2 high-noise. \emph{TA} / \emph{MQ} are averages over six text-alignment / seven motion-quality sub-dimensions ($1$--$5$ each). \emph{mean} averages the three judges, \emph{vote} is the per-dimension majority (ties broken upward). \emph{Real video} is an oracle upper bound. Best non-oracle results are shown in \textbf{bold}, and the sub-dimension breakdown is provided in App.~\ref{sec:appendix:vlmrubric}.}
\label{tab:main}
\small
\begin{tabular}{lcccc}
\toprule
Method & TA mean & TA vote & MQ mean & MQ vote \\
\midrule
Real video (oracle)                          & 4.5857 & 4.6477 & 4.4314 & 4.5805 \\
\midrule
Pretrained Wan2.2                            & 3.9189 & 3.9263 & 3.8181 & 3.8772 \\
SFT                                          & 4.1209 & 4.1393 & 3.7840 & 3.8509 \\
VideoREPA~\citep{zhang2025videorepa}         & 4.1252 & 4.1540 & 3.8024 & 3.8650 \\
MoAlign~\citep{bhowmik2025moalign}           & 4.1272 & 4.1537 & 3.8015 & 3.8711 \\
\textbf{SARA (ours)}                         & \textbf{4.1543} & \textbf{4.1668} & \textbf{3.8516} & \textbf{3.9191} \\
\bottomrule
\end{tabular}
\end{table}

\subsection{VBench Results}
\label{sec:exp:vbench}

We further evaluate these approaches on VBench-1.0~\citep{huang2024vbench} and VBench-2.0~\citep{zheng2025vbench} suites with their standard prompts and official scorers. Table~\ref{tab:vbench_combined} reports the VBench-1.0 \emph{Semantic} aggregate, the VBench-2.0 dimension scores, and the VBench-2.0 \emph{Final} score, with per-task breakdowns in App.~\ref{sec:appendix:vbench1} and App.~\ref{sec:appendix:vbench2}.

Both VBench protocols agree with the VLM rubric: SARA has the best aggregate score on each, leading VBench-1.0 \emph{Semantic} by $+0.90$ over VideoREPA and VBench-2.0 \emph{Final} by $+0.38$ over MoAlign. Per-dimension scores are more diffuse, as expected for sub-tasks that span very different aspects of generation. The small \emph{Human Fidelity} drop shared by all continually-trained methods is structural to the matched protocol: each approach updates only the Wan2.2 high-noise expert while leaving the low-noise expert frozen. Since anatomical detail is rendered at the low-noise stage of Wan2.2's two-expert mixture-of-experts (MoE), any high-noise update shifts the intermediate-latent distribution the un-updated low-noise expert was trained against (App.~\ref{sec:appendix:limitations}). SARA still posts the smallest such drop, consistent with its text-conditioned routing delivering a more targeted high-noise update. Across all three protocols, SARA is the only method that wins every aggregate score.

\begin{table}[htbp]
\centering
\caption{\textbf{Public VBench-1.0 / 2.0 results} ($\%$, higher is better). Best per column in \textbf{bold}.}
\label{tab:vbench_combined}
\small
\setlength{\tabcolsep}{4pt}
\resizebox{\linewidth}{!}{%
\begin{tabular}{l|c|cccccc}
\toprule
       & \textbf{VBench-1.0} & \multicolumn{6}{c}{\textbf{VBench-2.0}} \\
\cmidrule(lr){2-2}\cmidrule(lr){3-8}
Method & Semantic        & Creativity     & Commonsense    & Controllability & Human Fidelity & Physics        & Final \\
\midrule
Pretrained Wan2.2                            & 72.74          & 52.56          & 58.50          & 30.98           & \textbf{86.04} & 46.89          & 55.00 \\
SFT                                          & 72.17          & 54.60          & 59.68          & 29.59           & 80.41          & \textbf{51.09} & 55.08 \\
VideoREPA~\citep{zhang2025videorepa}         & 72.99          & 54.08          & \textbf{61.11} & \textbf{31.54}  & 82.78          & 46.67          & 55.24 \\
MoAlign~\citep{bhowmik2025moalign}           & 72.95          & \textbf{56.75} & 59.67          & 30.08           & 84.75          & 47.82          & 55.81 \\
\textbf{SARA (ours)}                         & \textbf{73.89} & 55.38          & \textbf{61.11} & 30.91           & 85.07          & 48.50          & \textbf{56.19} \\
\bottomrule
\end{tabular}%
}
\end{table}

\subsection{User study}
\label{sec:exp:userstudy}
\begin{figure}[htbp]
\centering
\includegraphics[width=0.85\linewidth]{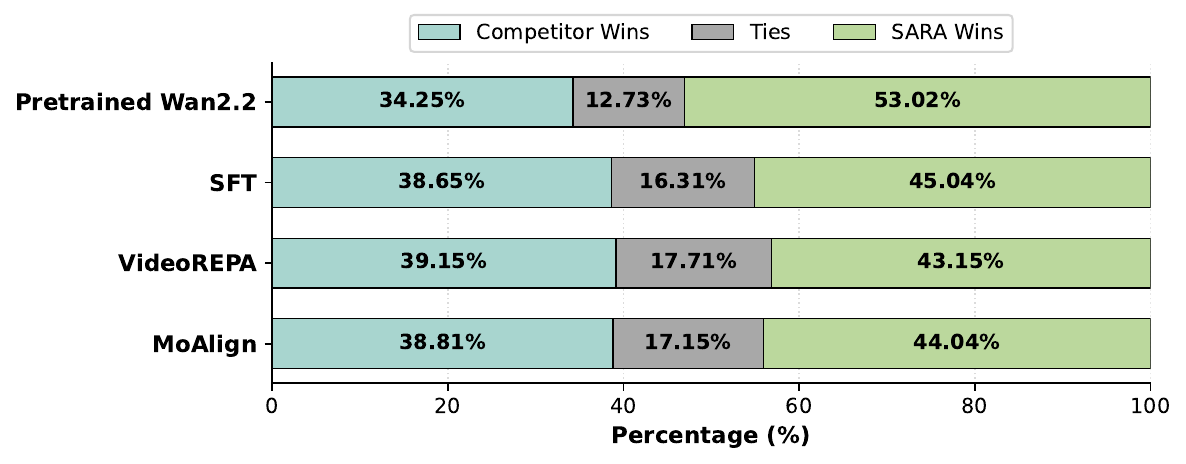}
\caption{\textbf{Blind pairwise user study}. Each row reports the percentage of comparisons where annotators prefer SARA, tie, or prefer the baseline. SARA is preferred over every baseline.}
\label{fig:userstudy}
\end{figure}

We also run a blind pairwise user study on a $200$-clip subset of the multi-entity test set, comparing SARA against the four baselines. For each pairing, annotators view side-by-side renderings of the same caption and pick a winner (or declare a tie). Fig.~\ref{fig:userstudy} shows SARA is preferred over all baselines, with the largest margin against the pretrained model and consistent gains over VideoREPA and MoAlign. The ordering aligns with the rubric-based and VBench results.

\subsection{Ablations}
\label{sec:exp:ablation}

Table~\ref{tab:ablation} ablates two components: the pair-routing operator of Eq.~\eqref{eq:pair-ops} (AND, OR, XOR) and the Stage~1 recipe (InfoNCE, entity-separated supervision, saliency head, temporal mask), plus MoAlign-style cross-frame decay ($\tau =10$) on top of SARA. Every variant keeps the main-run Stage~2 schedule and V-JEPA target, and only the indicated component is toggled. The pair-routing block reuses the SARA saliency aligner for all three operators.

\begin{table}[htbp!]
  \centering
  \caption{\textbf{SARA ablations} on the same Wan2.2 high-noise setup as Tab.~\ref{tab:main}, with each row toggling one design choice while the rest of SARA is fixed. Best in \textbf{bold}.}
  \label{tab:ablation}
  \small
  \begin{tabular}{lcccc}
  \toprule
  Variant & TA mean & TA vote & MQ mean & MQ vote \\
  \midrule
  \textbf{SARA (full, OR)}    & \textbf{4.1543} & \textbf{4.1668} & \textbf{3.8516} & \textbf{3.9191} \\
  \midrule
  \multicolumn{5}{l}{\emph{Pair-routing operator (Eq.~\eqref{eq:pair-ops})}} \\
  \quad XOR                   & 4.1107 & 4.1287 & 3.8043 & 3.8702 \\
  \quad AND                   & 4.1227 & 4.1532 & 3.8300 & 3.8995 \\
  \quad MoAlign (Tab.~\ref{tab:main})   & 4.1272 & 4.1537 & 3.8015 & 3.8711 \\
  \midrule
  \multicolumn{5}{l}{\emph{Saliency construction \& schedule}} \\
  \quad w/o InfoNCE                 & 4.1364 & 4.1575 & 3.8039 & 3.8721 \\
  \quad w/o entity-separated        & 4.1294 & 4.1587 & 3.8100 & 3.8693 \\
  \quad w/o saliency head           & 4.0775 & 4.0979 & 3.7851 & 3.8491 \\
  \quad w/o temporal mask           & 4.1405 & 4.1658 & 3.8153 & 3.8860 \\
  \quad w/ temporal decay $\tau=10$ & 4.1385 & 4.1572 & 3.8315 & 3.8981 \\
  \bottomrule
  \end{tabular}
  \end{table}

\paragraph{Saliency construction.} The saliency head is the largest single contributor: removing it and falling back to an NCE-only variant produces the largest drop on both TA and MQ. Ablating InfoNCE hurts both, with a larger drop on motion, consistent with the Stage~1 ablation (Fig.~\ref{fig:stage1_ablation}, Tab.~\ref{tab:stage1_diagnostics}) where \emph{w/o NCE} pushes mass onto the dominant subject, suppressing subject-background relations. Replacing the $K + 2$ entity-separated forwards with a single union-mask forward collapses $M_p$ into one foreground blob and hurts both dimensions too.

\paragraph{Pair-routing operator and temporal weighting.} Among the three operators in Eq.~\eqref{eq:pair-ops}, OR dominates: XOR keeps only FG-BG boundary pairs and drops the FG-FG structural relations, while AND trails OR because it discards FG-BG grounding. MoAlign induces an AND-like bias via its motion subspace and lands close to (saliency, AND) but below (saliency, OR). Removing the saliency mask from the cross-frame term (\emph{w/o temporal mask}) also trails full SARA, so the cross-frame term benefits from saliency-weighted pair selection. Adding MoAlign-style decay ($\tau{=}10$) on top of SARA does not help, confirming that gains come from saliency routing rather than temporal weighting.

These ablations pin SARA's gain to two components: a calibrated text-conditioned saliency in Stage~1, and an OR pair-routing operator in Stage~2 that keeps both subject-subject structure and subject-background grounding. Remove either one, or replace OR with AND/XOR, and SARA falls back towards the existing TRD baselines.

\subsection{Qualitative comparison}
\label{sec:exp:qualitative}

\begin{figure}[htbp]
\centering
\includegraphics[width=\linewidth]{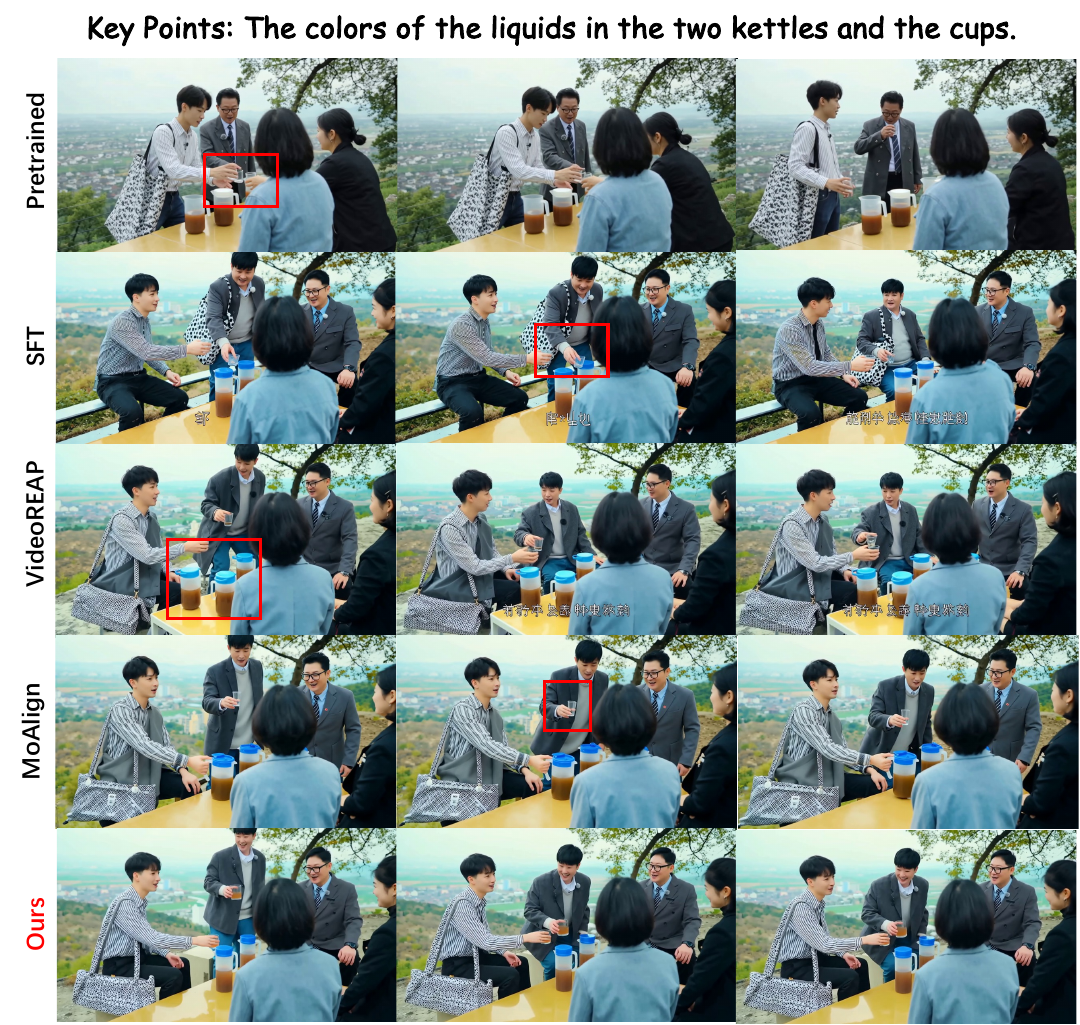}
\caption{\textbf{Qualitative comparison on fine-grained attribute binding.} The caption names distinct liquid colours for two kettles and two cups. Baselines mis-bind or wash out the colours (the failure mode of background-diluted alignment), whereas SARA renders each container with its specified colour. Matched-setting baselines: pretrained Wan2.2, SFT, VideoREPA, MoAlign.}
\label{fig:qual_case1}
\end{figure}

\begin{figure}[htbp]
\centering
\includegraphics[width=\linewidth]{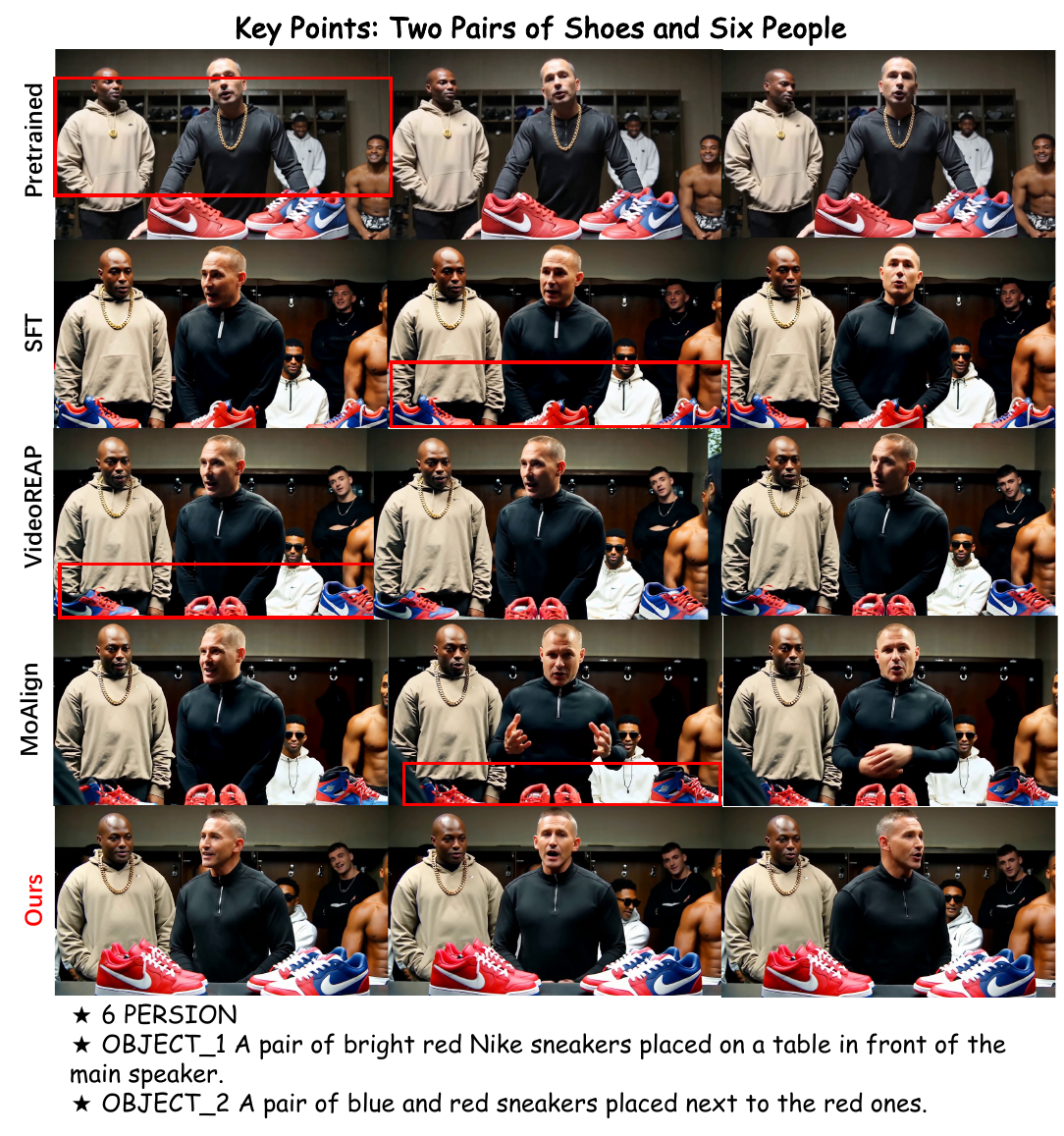}
\caption{\textbf{Qualitative comparison on a multi-entity scene} with six people and two pairs of sneakers (red Nike and blue-red). Baselines miss people, merge identities, or confuse the two shoe pairs, whereas SARA faithfully renders all six subjects and both correctly-coloured pairs. Matched-setting baselines: pretrained Wan2.2, SFT, VideoREPA, MoAlign.}
\label{fig:qual_case2}
\end{figure}

Direct visual inspection on two complementary failure modes corroborates the quantitative protocols, again against the four matched-setting baselines (pretrained Wan2.2, SFT, VideoREPA, MoAlign). Fig.~\ref{fig:qual_case1} isolates attribute binding: the caption names distinct liquid colours for two kettles and two cups, so the failure is a mis-routed attribute rather than a missing entity. The pretrained model and SFT swap or wash out the colours, VideoREPA and MoAlign recover only part of the binding, and SARA renders each container with its prompt-specified colour. Fig.~\ref{fig:qual_case2} stresses multi-entity coverage: a dense scene of six people and two pairs of sneakers, where baselines drop people, merge identities, or confuse the shoe pairs, while SARA recovers all six subjects and both pairs. Both cases match the pair-routing prediction: keeping subject--background pairs alongside subject--subject pairs preserves the relations that anchor an attribute or identity to the correct subject. Side-by-side video comparisons are on the project page.

\section{Conclusion}
\label{sec:conclusion}

SARA reframes semantic guidance for VDM representation alignment as a pair-routing problem on top of token-relation distillation. A lightweight Stage~1 aligner trained with per-entity SAM~3.1 masks and an InfoNCE regulariser predicts a continuous text-conditioned saliency, which is fused into TRD at Stage~2 through an OR pair-routing operator. This reallocates supervision from background-background pairs toward subject-subject and subject-background relations, while leaving the TRD form, V-JEPA target, and trainable diffusion-path parameter count unchanged. Under a matched experimental setup, SARA consistently outperforms SFT, VideoREPA, and MoAlign across all three evaluation protocols.


\appendix

\section{Pair-budget analysis}
\label{sec:appendix:pair_budget}

\begin{figure}[htbp]
\centering
\includegraphics[width=0.85\linewidth]{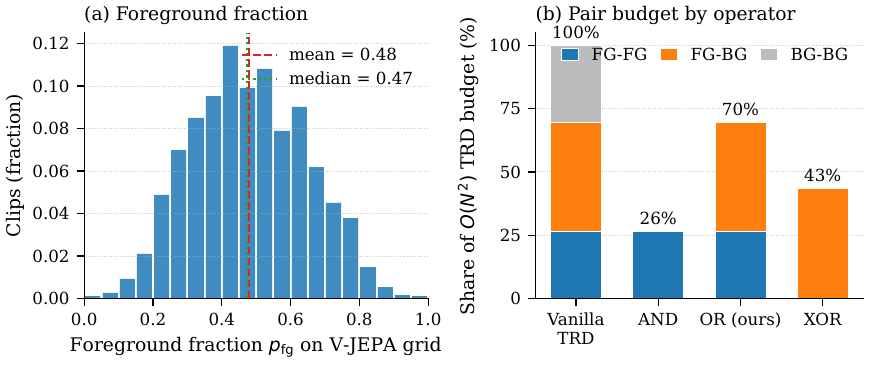}
\caption{\textbf{Pair-budget breakdown on the training corpus} ($2{,}400$ clips, $76{,}800$ frames). \textbf{(a)} Distribution of per-clip foreground fraction $p_\text{fg}$. \textbf{(b)} Share of the $O(N^2)$ TRD budget consumed by each pair category (FG--FG, FG--BG, BG--BG) under vanilla TRD (uniform weighting) and the three routing operators (AND, OR, XOR). OR (ours) retains $\sim\!70\%$ of pairs by keeping all FG--FG and FG--BG pairs and discarding only BG--BG.}
\label{fig:pair_budget}
\end{figure}

This appendix measures the routing gap that motivates SARA (Sec.~\ref{sec:method:motivation}). We compute the per-clip foreground fraction $p_\text{fg}$ from the per-entity SAM~3.1 masks that supervise Stage~1, over $2{,}400$ training clips ($76{,}800$ frames). Fig.~\ref{fig:pair_budget}(a) shows $p_\text{fg}$ concentrates around $0.48$, so slightly under half of a typical token grid is prompt-relevant foreground. Under uniform weighting the expected budget shares follow directly: $p_\text{fg}^2$ for subject--subject pairs, $2\,p_\text{fg}(1-p_\text{fg})$ for subject--background, and $(1-p_\text{fg})^2$ for background--background. Fig.~\ref{fig:pair_budget}(b) reports the realised split: background--background pairs the caption never references consume roughly $30\%$ of the supervision while subject--subject relations receive only $\sim\!26\%$. SARA's OR operator reclaims this $\sim\!30\%$ by keeping every FG--FG and FG--BG pair and discarding only BG--BG, the reallocation behind the gains in Tab.~\ref{tab:ablation}.

\section{Stage-1 saliency aligner: supervision, ablations, and diagnostics}
\label{sec:appendix:viz}

This appendix presents the evidence behind the Stage-1 design of Sec.~\ref{sec:method:stage1}: the SAM~3.1 entity decomposition that forms the supervision target (App.~\ref{sec:appendix:viz_sam3}), and an ablation of the two routing-critical ingredients, the entity-separated $K+2$ supervision and the InfoNCE regulariser, with a quantitative diagnostic for each failure mode (App.~\ref{sec:appendix:viz_ablation}). The trained aligner's per-query and full-caption behaviour is shown in the main text (Figs.~\ref{fig:viz_stage1}--\ref{fig:viz_stage2}).

\subsection{SAM~3.1 entity decomposition (supervision target)}
\label{sec:appendix:viz_sam3}

\begin{figure}[htbp]
\centering
\includegraphics[width=0.82\linewidth]{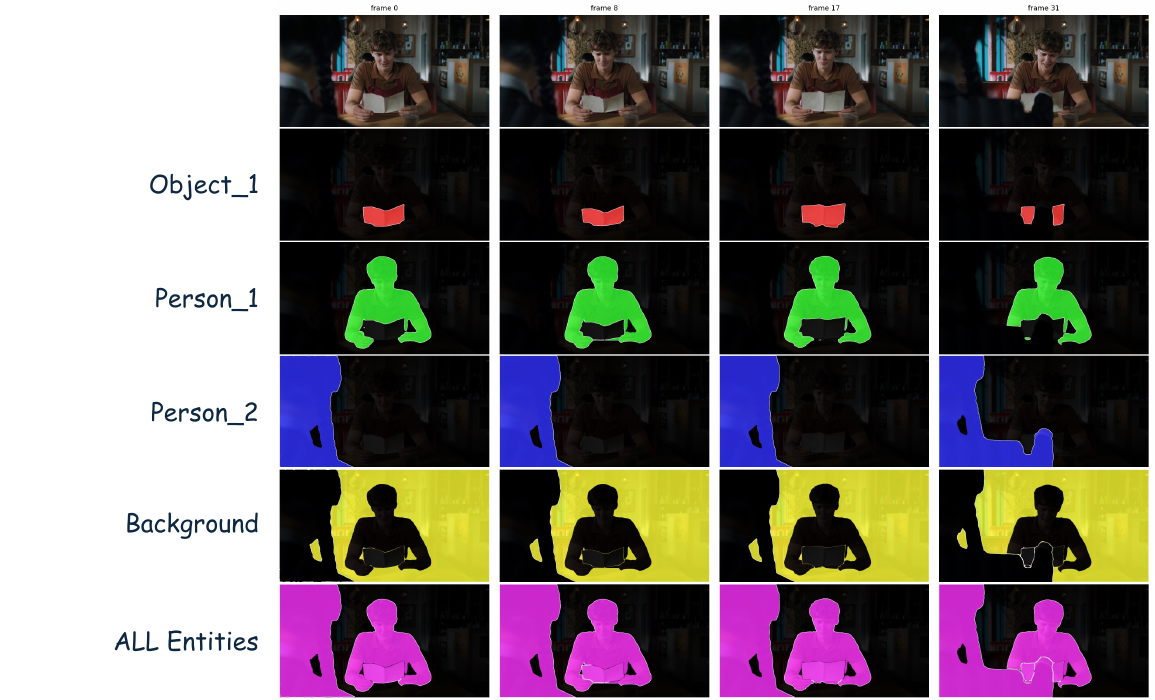}
\caption{\textbf{SAM~3.1 entity decomposition used as Stage~1 supervision.} Top row: input frames. Next three rows: per-entity masks for \texttt{OBJECT\_1} (red), \texttt{PERSON\_1} (green), \texttt{PERSON\_2} (blue). Last two rows: complement \texttt{BACKGROUND} ($M_\text{bg} =\mathbf{1} - M_\text{fg}$, yellow) and foreground union \texttt{ALL Entities} ($M_\text{fg}$, magenta). All five masks supervise the saliency head jointly via $K + 2$ forwards (Sec.~\ref{sec:method:stage1}).}
\label{fig:sam3_entity_masks}
\end{figure}

Fig.~\ref{fig:sam3_entity_masks} unpacks one training clip into the five binary masks Stage~1 supervises against (Sec.~\ref{sec:method:stage1}): one per-entity mask $M_k$ per Reference item, the foreground union $M_\text{fg} =\bigcup_k M_k$, and the complement $M_\text{bg} =\mathbf{1} -M_\text{fg}$, all produced offline by the SAM~3.1 pipeline of App.~\ref{sec:appendix:simplifier}. The clip exposes two properties that shape the saliency design. First, the foreground entities span very different scales (a small held card against a partially off-frame person), so a single union-mask forward would let the dominant entity erase the small ones, the motivation for the $K+2$ entity-separated forwards. Second, \texttt{PERSON\_2}'s mask tracks the body even where it leaves the frame, so SAM~3.1's open-vocabulary prompting recovers named entities under partial framing. Together they keep the entity-separated supervision well-defined on the crowded multi-subject clips SARA targets.

\subsection{Ablation grid and quantitative diagnostics}
\label{sec:appendix:viz_ablation}

\begin{figure}[htbp]
\centering
\begin{subfigure}[t]{0.95\linewidth}\centering
  \includegraphics[width=\linewidth]{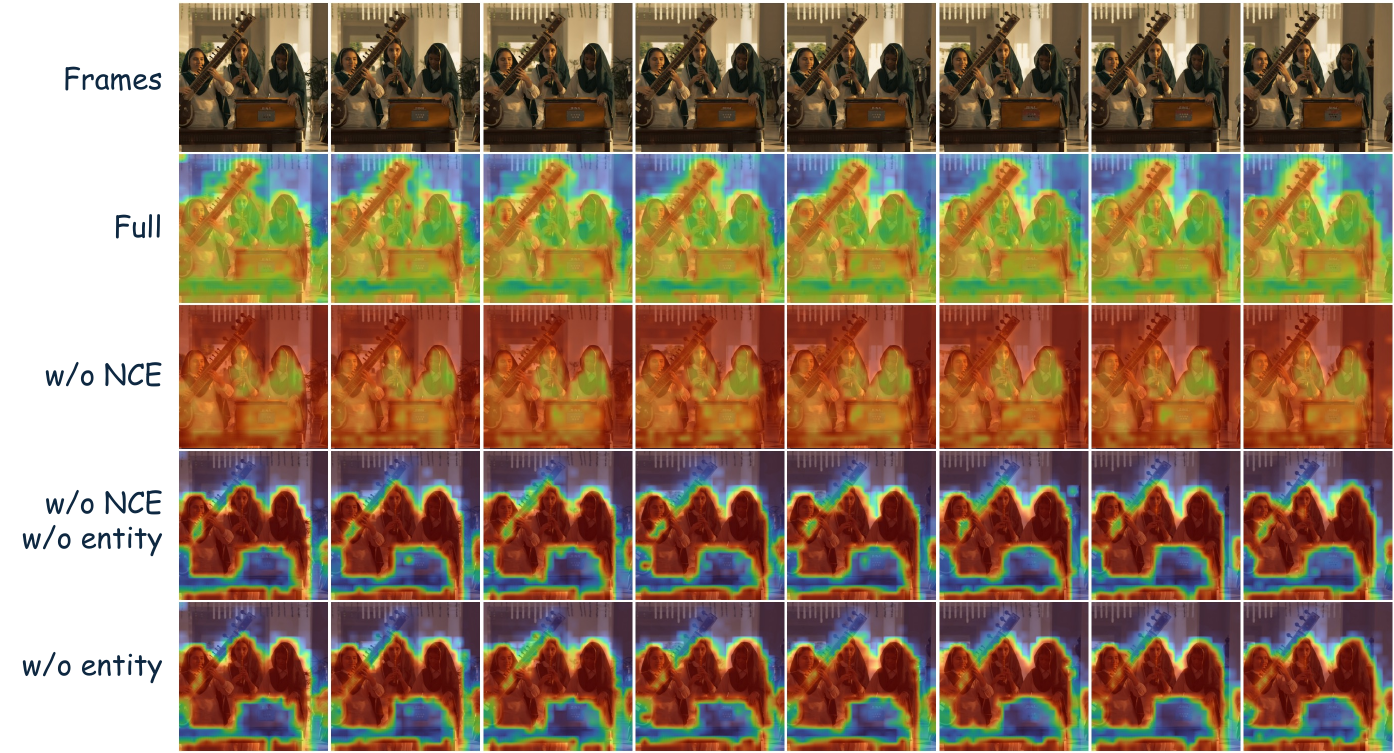}
  \caption{Clip A: South Asian musicians playing instruments.}
  \label{fig:stage1_ablation:case1}
\end{subfigure}\\[0.5em]
\begin{subfigure}[t]{0.95\linewidth}\centering
  \includegraphics[width=\linewidth]{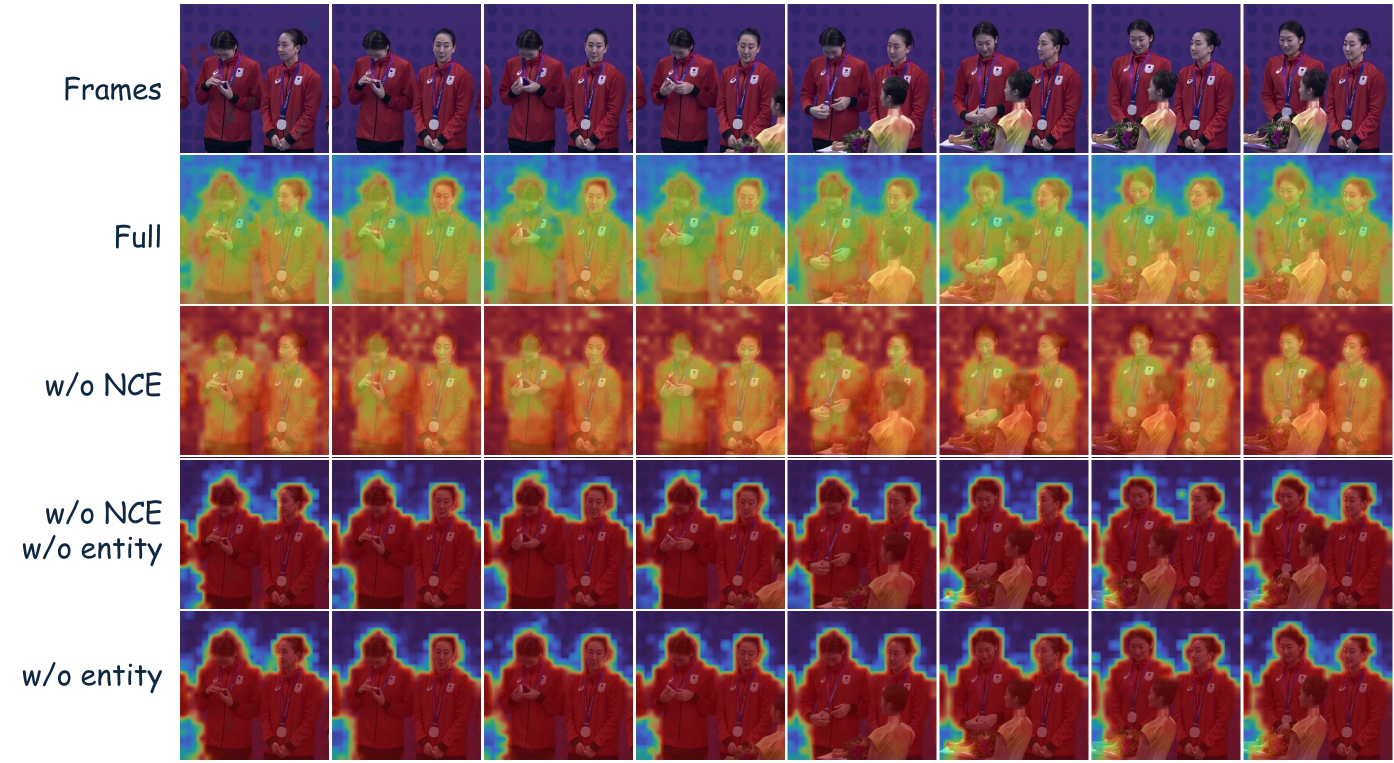}
  \caption{Clip B: women in red ceremonial attire posing as a tightly grouped multi-subject foreground.}
  \label{fig:stage1_ablation:case2}
\end{subfigure}
\caption{\textbf{Stage~1 ablation on the predicted saliency map} on two held-out clips, eight frames each. Rows: input frames; \emph{Full} (default SARA, $K + 2$ forwards $+$ InfoNCE); \emph{w/o NCE}; \emph{w/o NCE w/o entity} (single forward on full caption with union mask, no InfoNCE); \emph{w/o entity} (single forward, InfoNCE retained). Saliency rows use a jet colormap: redder = higher saliency, bluer = lower. Discussion in App.~\ref{sec:appendix:viz_ablation}, quantitative metrics in Tab.~\ref{tab:stage1_diagnostics}.}
\label{fig:stage1_ablation}
\end{figure}

Fig.~\ref{fig:stage1_ablation} ablates the predicted saliency $M_p$ along the two routing-critical Stage~1 choices (Sec.~\ref{sec:method:stage1}; downstream results in Sec.~\ref{sec:exp:ablation}): the entity-separated $K + 2$ supervision and the InfoNCE regulariser, on two held-out clips with distinct layouts. Both \emph{w/o entity} rows saturate across nearly every patch, merging subjects into one foreground blob, the collapse that motivates entity-separated supervision. \emph{w/o NCE} is sharper but biased toward the dominant subject (the central musician in panel~(a), the front-row women in panel~(b)) at the expense of smaller entities. Only \emph{Full} produces a calibrated, per-entity response. Tab.~\ref{tab:stage1_diagnostics} quantifies these trends with seven metrics in four groups: saliency calibration, cross-attention focus, self-attention entropy gain, and rank preservation.

\paragraph{Quantitative metrics: definitions and rationale.}\label{para:diagnostic_defs}
Every entry of Tab.~\ref{tab:stage1_diagnostics} is a scalar averaged over a held-out set of $N = 128$ clips ($8$ frames each, disjoint from training). All quantities below are defined on a single (clip, frame):
\begin{itemize}\setlength{\itemsep}{1pt}
\item $M_p\in[0,1]^{N_v}$: predicted saliency over the $N_v$ V-JEPA patches; $M_p(n)$ is its value at patch $n$.
\item $\bar\Phi_\text{CA}\in\mathbb{R}^{N_v\times L}$: head-averaged cross-attention; the row $\bar\Phi_\text{CA}(n,\cdot)\in\mathbb{R}^{L}$ is patch $n$'s attention distribution over the $L$ text tokens.
\item $A_\text{vj},A_\text{post}\in\mathbb{R}^{N_h\times N_v\times N_v}$: visual self-attention of the $N_h$ heads, before (raw V-JEPA) and after the cross-attention stack; $A^{(h)}_{n,\cdot}$ is head $h$'s attention from patch $n$.
\item $V_y$, $V'_y=\Phi_\text{CA}(V_y,\tilde E)$: the raw and text-enhanced V-JEPA features of Sec.~\ref{sec:method:stage1}.
\end{itemize}
Two entropies recur: the Shannon entropy $H(p)=-\sum_j p_j\log p_j$ of a probability vector $p$, and the Bernoulli entropy $H_2(q)=-q\log q-(1-q)\log(1-q)$ of a scalar $q\in[0,1]$. The indicator $\mathbf{1}[\cdot]$ is $\{0,1\}$-valued. The seven metrics fall into four groups, each targeting one failure mode of Fig.~\ref{fig:stage1_ablation}.

\textbf{Saliency calibration} (rows 1--4; foreground-prior collapse and over-binarisation).
\begin{itemize}\setlength{\itemsep}{1pt}
\item saliency mean, $\tfrac{1}{N_v}\sum_n M_p(n)$: average firing level. Values near the foreground prior ($\approx 0.6$ for our $K = 2$ entities) mean the head fires almost everywhere.
\item saliency max, $\max_n M_p(n)$: a peak near $1$ means the sigmoid has saturated and no longer outputs a graded signal.
\item saliency coverage, $\tfrac{1}{N_v}\sum_n \mathbf{1}[M_p(n)>0.5]$: fraction of patches above threshold; smaller is more selective.
\item saliency entropy, $\tfrac{1}{N_v}\sum_n H_2(M_p(n))$: high values keep $M_p$ graded near $0.5$, the regime the OR weight $W^{\vee}$ of Eq.~\eqref{eq:pair-ops} consumes; low values mean $M_p$ has hardened into a $\{0,1\}$ mask and $W^{\vee}$ degenerates to an indicator.
\end{itemize}

\textbf{Cross-attention focus} (row 5; whether CA reads the caption).
\begin{itemize}\setlength{\itemsep}{1pt}
\item ca-focus mean, $\tfrac{1}{N_v}\sum_n\bigl(1-H(\bar\Phi_\text{CA}(n,\cdot))/\log(L+1)\bigr)$: one minus the normalised text-attention entropy of each patch. Higher means a patch attends to a few specific words rather than spreading uniformly, so the head can route by word identity.
\end{itemize}

\textbf{Self-attention entropy gain} (row 6; whether CA enriches or collapses V-JEPA self-attention).
\begin{itemize}\setlength{\itemsep}{1pt}
\item $\Delta$ self-attn entropy, $\bar H(A_\text{post})-\bar H(A_\text{vj})$, where $\bar H(A)=\tfrac{1}{N_h N_v}\sum_{h,n} H(A^{(h)}_{n,\cdot})/\log N_v$ is the length-normalised Shannon entropy averaged over heads and patches ($\bar H(A_\text{vj})=0.783$ here). A positive value means CA adds spread on top of V-JEPA; a negative value (w/o NCE w/o entity: $-0.085$) means CA narrows attention onto a single foreground blob.
\end{itemize}

\textbf{Representation-rank preservation} (row 7; whether CA keeps V-JEPA's high-rank structure).
\begin{itemize}\setlength{\itemsep}{1pt}
\item PCA $\Delta$ var.\ ratio, $r_3(V'_y)-r_3(V_y)$, with $r_3(V)=\sum_{i\le 3}\sigma_i^2/\sum_i\sigma_i^2$ the top-3 explained-variance ratio and $\sigma_i$ the $i$-th singular value of the centred token matrix. A small positive value keeps $V'_y$ within V-JEPA's rank profile; a large one means CA has compressed the tokens into a low-rank, foreground-only subspace.
\end{itemize}

The seven rows form a conjunction: only Full avoids saturation (rows 1--3), keeps $M_p$ graded (row 4), focuses CA on entity-specific words (row 5), widens rather than narrows self-attention (row 6), and preserves the rank of $V_y$ (row 7); each ablation breaks at least one.

\begin{table}[htbp]
\centering
\small
\caption{\textbf{Stage~1 ablation: quantitative metrics} (mean over $N = 128$ held-out clips, columns match the rows of Fig.~\ref{fig:stage1_ablation}, metric definitions on p.\,\pageref{para:diagnostic_defs}). \emph{Full} is the only configuration that jointly avoids saliency saturation (rows~1--3), keeps $M_p$ graded (row~4), focuses cross-attention on entity tokens (row~5), and enriches V-JEPA self-attention without collapsing it (rows~6--7), and each ablation fails on at least one metric.}
\label{tab:stage1_diagnostics}
\begin{tabular}{lcccc}
\toprule
Metric & \emph{Full} & \emph{w/o NCE} & \emph{w/o NCE w/o entity} & \emph{w/o entity} \\
\midrule
saliency mean        & 0.407 & 0.567 & 0.653 & 0.648 \\
saliency max         & 0.865 & 0.925 & 0.996 & 0.998 \\
saliency coverage    & 0.383 & 0.606 & 0.657 & 0.658 \\
saliency entropy     & 0.463 & 0.461 & 0.109 & 0.207 \\
ca-focus mean        & 0.221 & 0.159 & 0.068 & 0.118 \\
$\Delta$ self-attn entropy & $+0.065$ & $+0.096$ & $-0.085$ & $+0.013$ \\
PCA $\Delta$ var.\ ratio & $+0.083$ & $+0.164$ & $+0.260$ & $+0.183$ \\
\bottomrule
\end{tabular}
\end{table}

\section{MTSS captioning and entity-mask preparation}
\label{sec:appendix:mtss}

\subsection{MTSS caption format and pipeline}
\label{sec:appendix:mtss:format}

\paragraph{Format.} Stage~2 inference needs a global, video-level caption for both the VDM and the frozen aligner. We adopt the MTSS format~\citep{team2026script}, which factorises a video into four streams (\textbf{Reference} for persistent entities and scenes, \textbf{Shot} for visual segments, \textbf{Event} for localised audio/interaction events, and \textbf{Global} for ambient context) linked by stable \texttt{ref\_id}s (e.g.\ \texttt{PERSON\_1}, \texttt{OBJECT\_1}, \texttt{SCENE\_1}) and per-shot \texttt{time\_range}s. The Reference stream gives ready-made per-entity captions $c_k$ for Stage~1 supervision, \texttt{ref\_id} dereferencing keeps the full caption under Qwen3-VL-Embedding's $2048$-token cap, and the stream-level separation reduces the foreground/background entanglement that entity-separated Stage~1 training exploits. Listing~\ref{lst:mtss} shows a compact example of the resulting JSON.

\begin{lstlisting}[caption={A compact MTSS JSON example. Persistent entities live under \texttt{references} with stable \texttt{ref\_id}s and are referenced from each \texttt{shots[i].visual\_description} and \texttt{references\_in\_shot}, so per-entity captions $c_k$ and per-shot timing are read off the same structure.}, label={lst:mtss}]
{
  "structured_caption": { "english": {
    "scene_description": "In an office, a woman in a red dress argues on the phone.",
    "global_style": "Realistic HD; bright office lighting; tense pace.",

    "references": [
      { "ref_id": "PERSON_1", "type": "person",
        "semantic_description": "A young woman intensely arguing on the phone.",
        "appearance_anchor": {
          "id_features": { "detail_description":
            "East Asian, late 20s; long wavy dark-brown hair; red lipstick." },
          "attributes": {
            "clothing":    "Fitted V-neck mini dress in vibrant red.",
            "hairstyle":   "Long, wavy dark-brown hair worn down.",
            "accessories": "Light jade bracelet on the left wrist." } } },

      { "ref_id": "OBJECT_1", "type": "object",
        "semantic_description": "The smartphone the woman is using.",
        "appearance_anchor": { "id_features": { "detail_description":
          "Black smartphone in a black case, held to her right ear." } } },

      { "ref_id": "SCENE_1", "type": "scene",
        "semantic_description": "A modern office interior.",
        "appearance_anchor": { "id_features": { "detail_description":
          "Large dark-grey desk with hardcover books; bookshelf and beige chair behind." } } }
    ],

    "shots": [
      { "shot_id": "shot_1", "time_range": [0.0, 3.4],
        "references_in_shot": ["PERSON_1", "OBJECT_1", "SCENE_1"],
        "camera": { "movement": "static", "angle": "eye-level", "shot_type": "medium" },
        "visual_description":
          "In SCENE_1, PERSON_1 leans over her desk holding OBJECT_1 to her right ear. Her expression shifts from tense concentration to deep anger; she bares her teeth in a snarl while glaring off-camera." }
    ]
  } }
}
\end{lstlisting}

\paragraph{Pipeline.} For each training video $x_0$ we obtain MTSS captions in three offline steps, without human annotation. (i)~The dense narrative caption provided with each training clip and $32$ uniformly sampled frames are fed to Qwen3-VL-72B in vision-language mode; the system prompt instructs the model to enumerate Reference items with \texttt{ref\_id}s and short \texttt{semantic\_description}s, segment the video into Shots, extract Events with \texttt{time\_range}s, and write a single Global summary, serialised in the MTSS format. (ii)~Each foreground Reference item (\texttt{PERSON\_*}/\texttt{OBJECT\_*}) yields an entity caption $c_k$ (the concatenation of its \texttt{semantic\_description} and \texttt{detail\_description} fields) and a SAM~3.1 mask $M_k$ from a Qwen3.5-2B-simplified noun phrase passed as SAM~3.1's text prompt (App.~\ref{sec:appendix:simplifier}); the background caption $c_\text{bg}$ is the \texttt{SCENE\_*}/\texttt{BACKGROUND} item's \texttt{semantic\_description}, with mask $M_\text{bg} = \mathbf{1} - M_\text{fg}$. (iii)~For Stage~2 inference and evaluation, captions exceeding $2$K tokens are compressed by Qwen3.5 while preserving all \texttt{ref\_id}s and \texttt{time\_range}s; Stage~1 reads \texttt{semantic\_description}s directly from the JSON and is unaffected. A flat user prompt at inference time is rewritten into MTSS form offline by the same pipeline. Rewriter limitations are discussed in App.~\ref{sec:appendix:limitations}.

\subsection{Linking MTSS entities to SAM~3.1 masks}
\label{sec:appendix:simplifier}

Stage~1 mask supervision relies on per-entity binary masks from SAM~3.1 Multiplex, and two practical issues make the raw MTSS-to-SAM pipeline unreliable. First, SAM~3.1 expects short noun-phrase prompts, whereas MTSS Reference items carry rich free-form descriptions. Second, when several same-type entities co-occur (e.g.\ two \texttt{PERSON\_*}), SAM~3.1 returns multiple instances without telling us which instance matches which Reference id. We address both with a small frozen Qwen3.5-2B model used in two complementary modes.

\paragraph{Text simplification.} Each MTSS entity description is rewritten by Qwen3.5-2B into a $2$--$5$-word noun phrase that keeps the most visually distinctive adjective(s). Examples taken from training logs:
\begin{itemize}\setlength\itemsep{0.1em}
    \item ``A young Caucasian male with fair skin, short curly light brown hair, wearing a navy hoodie'' $\to$ ``young curly-haired man''.
    \item ``A folded greeting card being held by the barista'' $\to$ ``folded card''.
    \item ``A dark wooden bar counter with several espresso cups arranged on top'' $\to$ ``wooden bar counter''.
\end{itemize}
The simplified phrase is passed to SAM~3.1 as the text prompt for that entity, which returns far more non-empty, well-covering masks than the raw description does.

\paragraph{Bounding-box detection for instance disambiguation.} For multi-instance types we additionally query Qwen3.5-2B in vision-language mode on the first video frame: given the original entity description and the frame, the model emits a bounding box. SAM~3.1 instance boxes are then matched to entity boxes by IoU, which assigns each Reference id to a single SAM~3.1 instance and hence to a per-entity mask. When no valid Qwen box is available we fall back to area-ranked assignment, which never drops below the SAM-only baseline.

\paragraph{Scope.} The simplifier is part of Stage~1 mask preparation only. Stage~2 does not invoke it, and inference is unchanged. Fig.~\ref{fig:sam3_entity_masks} (App.~\ref{sec:appendix:viz_sam3}) visualises the resulting per-entity mask decomposition on a representative training clip and motivates the entity-separated supervision adopted in Sec.~\ref{sec:method:stage1}.

\section{VLM-rubric evaluation protocol}
\label{sec:appendix:vlmrubric}

\paragraph{Judging setup.} For every generated video and each of the three VLM judges, we issue one chat completion per rubric (TA and MQ). The video and the MTSS caption (App.~\ref{sec:appendix:mtss}) are sent in a single multimodal turn with thinking mode enabled under an $8$K-token budget, so the judge reasons over the rubric and then emits a strict JSON object (Box~\ref{box:json:ta} / Box~\ref{box:json:mq}). Sampling follows each family's official thinking-mode recipe: Qwen3.x uses $T = 1.0$, top-$p = 0.95$, top-$k = 20$, $\min$-$p = 0$, presence penalty $1.5$, while Gemma-4 uses $T = 1.0$, top-$p = 0.95$, top-$k = 64$. The \emph{vote} aggregation resolves ties toward the higher score.

\paragraph{Protocol noise floor and the real-video oracle.} The \emph{Real video} row of Table~\ref{tab:main} sits clearly above every continually-trained method but below the rubric's $5.0$ ceiling. The sub-$5$ gap has two protocol-level sources independent of the generation pipeline: the MTSS caption is produced by an external VLM (App.~\ref{sec:appendix:mtss}) and occasionally hallucinates entities or mis-binds attributes, against which even the source video cannot satisfy the strict rubric; and the judge VLMs over-penalise small attribute mismatches, mis-count entities under occlusion, or down-score brief actions on \texttt{action\_completion}. Both apply to every row of Table~\ref{tab:main} and form a constant noise floor, so the oracle row should be read as the achievable protocol score; the relevant quantity is the gap each method closes toward it under matched data, schedule, judges, and captions. SARA closes the largest fraction of that gap on both TA and MQ.

\paragraph{Prompt template.} The prompt fed to each judge concatenates (i) the system prompt of Box~\ref{box:sys}, (ii) a \texttt{TEXT DESCRIPTION} block populated from the MTSS \texttt{[Scene \& Style]}, \texttt{[Characters \& Objects]} (with expected entity counts) and \texttt{[Shot Narrative]} sections, and (iii) one of the two rubric blocks (Box~\ref{box:rubric:ta} for alignment, Box~\ref{box:rubric:mq} for motion). The judge replies with a JSON object keyed by the rubric dimensions and valued by $\{\text{score},\text{reason}\}$ pairs (Box~\ref{box:json:ta} / Box~\ref{box:json:mq}).

\begin{tcolorbox}[promptbox, title={Box \promptlabel{box:sys}: System prompt (shared by both rubrics)}]
\small\ttfamily
You are an expert video evaluation assistant. You compare generated videos against their text descriptions and score alignment across multiple dimensions. Be strict and precise -- only give high scores when the video truly matches the description.
\end{tcolorbox}

\begin{tcolorbox}[promptbox, title={Box \promptlabel{box:rubric:ta}: Alignment rubric (TA, six dimensions, $1$--$5$ each)}]
\small\ttfamily
=== EVALUATION DIMENSIONS ===\\[2pt]
Score each ALIGNMENT dimension from 1 to 5 on a shared Likert scale: \textbf{5}=fully matches the description; \textbf{4}=mostly matches with minor differences; \textbf{3}=partially matches or some items clearly wrong; \textbf{2}=mostly wrong; \textbf{1}=does not match / unrecognizable. The only exception is \texttt{entity\_count}, which uses an exact-count scale: 5=exact match, 4=off by 1 total, 3=off by 2, 2=off by 3--4, 1=off by 5+.\\[4pt]
1.~\textbf{entity\_count}: Does the video contain the correct number of people (\{num\_persons\}) and objects (\{num\_objects\})?\\[2pt]
2.~\textbf{person\_appearance}: Do the people's clothing, hairstyle, age/gender, and accessories match the [Characters \& Objects] description?\\[2pt]
3.~\textbf{object\_appearance}: Do the objects' shape, color, material, and type match the [Characters \& Objects] description?\\[2pt]
4.~\textbf{spatial\_arrangement}: Are people and objects positioned as described in the [Shot Narrative] (left/right/center/between/behind, etc.)?\\[2pt]
5.~\textbf{action\_completion}: Are the actions and movements described in the [Shot Narrative] actually performed in the video?\\[2pt]
6.~\textbf{scene\_style}: Does the video's setting, lighting, color palette, and mood match the [Scene \& Style] description?
\end{tcolorbox}

\begin{tcolorbox}[promptbox, title={Box \promptlabel{box:json:ta}: Required JSON output (alignment)}]
\small\ttfamily
=== OUTPUT FORMAT ===\\[2pt]
Respond in JSON format ONLY (no extra text):\\[2pt]
\{\\
\hspace*{1em}"entity\_count":~~~~~~~~~~\{"score": <1-5>, "reason": "<brief reason>"\},\\
\hspace*{1em}"person\_appearance":~~~~~\{"score": <1-5>, "reason": "<brief reason>"\},\\
\hspace*{1em}"object\_appearance":~~~~~\{"score": <1-5>, "reason": "<brief reason>"\},\\
\hspace*{1em}"spatial\_arrangement":~~~\{"score": <1-5>, "reason": "<brief reason>"\},\\
\hspace*{1em}"action\_completion":~~~~~\{"score": <1-5>, "reason": "<brief reason>"\},\\
\hspace*{1em}"scene\_style":~~~~~~~~~~~\{"score": <1-5>, "reason": "<brief reason>"\}\\
\}
\end{tcolorbox}

\begin{tcolorbox}[promptbox, title={Box \promptlabel{box:rubric:mq}: Motion-quality rubric (MQ, seven dimensions, $1$--$5$ each)}]
\small\ttfamily
=== EVALUATION DIMENSIONS ===\\[2pt]
Score each MOTION dimension from 1 to 5 on a shared Likert scale: \textbf{5}=requirement fully met (or, where applicable, the described motion happens with correct subject/object/direction/timing); \textbf{4}=mostly met with minor attribute or magnitude issues; \textbf{3}=partially met with clear localized failures or mismatches; \textbf{2}=multiple or large-scale failures, only a vague hint of the requirement; \textbf{1}=requirement fails entirely, or behaves opposite to the description.\\[4pt]
1.~\textbf{motion\_prompt\_alignment}: Do the actions described in the [Shot Narrative] (subject + verb + object + direction) actually happen in the video?\\[2pt]
2.~\textbf{motion\_completeness}: Does each motion have a coherent start $\to$ middle $\to$ end trajectory (clear onset, full execution, clean closure)? Penalize twitching, frozen frames, and on-the-spot looping.\\[2pt]
3.~\textbf{motion\_amplitude}: Is the amount/frequency of motion reasonable for the described scene? Penalize pseudo-static clips (only micro-jitter pretending to be motion) and over-exaggerated shaking, seizure-like flicker, or cartoon distortion.\\[2pt]
4.~\textbf{temporal\_consistency}: Do subjects keep their identity across frames? Treat as defects: flicker, ghosting, teleportation, identity swap, extra/missing fingers/limbs, limb displacement or duplication, facial scrambling, edge melting/tearing, color bleeding.\\[2pt]
5.~\textbf{physical\_plausibility}: Does the motion respect physics (gravity, inertia, collision, rigid-vs-soft body, contact)? Treat as defects: joints twisted past anatomical limits, rigid objects wobbling like cloth, limbs/objects clipping through solid surfaces, feet floating or sinking, fluid/smoke moving against gravity.\\[2pt]
6.~\textbf{camera\_motion}: Does the camera behave as described? If the [Shot Narrative] specifies a move (push-in / pull-out / pan / tilt / tracking / orbit / handheld), check type, direction, and pace, otherwise expect a stable camera.\\[2pt]
7.~\textbf{interaction\_correctness}: Do multi-subject interactions in the [Shot Narrative] (physical contact, coordination, mutual effect) actually happen with correct contact point and approach$\to$contact$\to$completion timing on both sides?\\
\hspace*{1em}NOTE: if the caption involves only a single subject, return 5 with reason ``caption has no multi-subject interaction''.
\end{tcolorbox}

\begin{tcolorbox}[promptbox, title={Box \promptlabel{box:json:mq}: Required JSON output (motion)}]
\small\ttfamily
Respond in JSON format ONLY (no extra text):\\[2pt]
\{\\
\hspace*{1em}"motion\_prompt\_alignment":~~~\{"score": <1-5>, "reason": "<brief reason>"\},\\
\hspace*{1em}"motion\_completeness":~~~~~~~\{"score": <1-5>, "reason": "<brief reason>"\},\\
\hspace*{1em}"motion\_amplitude":~~~~~~~~~~\{"score": <1-5>, "reason": "<brief reason>"\},\\
\hspace*{1em}"temporal\_consistency":~~~~~~\{"score": <1-5>, "reason": "<brief reason>"\},\\
\hspace*{1em}"physical\_plausibility":~~~~~\{"score": <1-5>, "reason": "<brief reason>"\},\\
\hspace*{1em}"camera\_motion":~~~~~~~~~~~~~\{"score": <1-5>, "reason": "<brief reason>"\},\\
\hspace*{1em}"interaction\_correctness":~~~\{"score": <1-5>, "reason": "<brief reason>"\}\\
\}
\end{tcolorbox}

\paragraph{Per-sub-dimension scores.} Tabs.~\ref{tab:appendix:align-dims} and~\ref{tab:appendix:motion-dims} report the mean-aggregated per-sub-dimension scores for the methods of Tab.~\ref{tab:main} and the ablations of Tab.~\ref{tab:ablation}. On the alignment side the bottleneck is \texttt{action\_completion}: every continually-trained method sits well below the real-video oracle there, making fine-grained action coverage the hardest TA dimension, with SARA still posting the largest gain among non-oracles. On the motion side the gains concentrate on \texttt{motion\_prompt\_alignment}, \texttt{motion\_completeness}, \texttt{motion\_amplitude}, and \texttt{interaction\_correctness}, while the conservative pretrained baseline keeps a small lead on \texttt{temporal\_consistency} and \texttt{physical\_plausibility}. These two dimensions share the same mechanistic origin as the VBench-2.0 \emph{Human Fidelity} drop discussed in Sec.~\ref{sec:exp:vbench}: temporal consistency, physical plausibility, and anatomical fidelity are all rendered at the low-noise stage of Wan2.2's two-expert MoE, but our matched protocol updates only the high-noise expert, so high-noise updates that improve coarse-structure prompt following propagate as a small distribution shift on the un-updated low-noise expert (App.~\ref{sec:appendix:limitations}). SARA shows the smallest such drop among the three continually-trained methods.

\begin{table}[htbp]
\centering
\caption{\textbf{Per-sub-dimension alignment scores (mean across three VLM judges).} Best non-oracle in \textbf{bold}. \emph{EntCnt}=entity\_count, \emph{PrsApp}=person\_appearance, \emph{ObjApp}=object\_appearance, \emph{Spatial}=spatial\_arrangement, \emph{ActCmp}=action\_completion, \emph{Scene}=scene\_style.}
\label{tab:appendix:align-dims}
\footnotesize
\begin{tabular}{lccccccc}
\toprule
Method & EntCnt & PrsApp & ObjApp & Spatial & ActCmp & Scene & Avg \\
\midrule
\multicolumn{8}{l}{\emph{Main comparison}} \\
\midrule
Real video (oracle)         & 4.299 & 4.579 & 4.605 & 4.708 & 4.340 & 4.983 & 4.586 \\
\midrule
Pretrained Wan2.2           & 3.930 & 3.654 & 4.461 & 3.700 & 2.863 & 4.905 & 3.919 \\
SFT                         & 4.276 & 4.021 & \textbf{4.515} & 3.951 & 3.048 & 4.913 & 4.121 \\
VideoREPA                   & 4.298 & 4.030 & 4.489 & 3.974 & 3.050 & 4.909 & 4.125 \\
MoAlign                     & 4.282 & 4.048 & 4.504 & 3.955 & 3.053 & 4.921 & 4.127 \\
\textbf{SARA (ours)}        & \textbf{4.301} & \textbf{4.085} & 4.502 & \textbf{4.002} & \textbf{3.114} & \textbf{4.923} & \textbf{4.154} \\
\midrule
\multicolumn{8}{l}{\emph{Ablations}} \\
\midrule
\textbf{SARA (full)}        & 4.301 & \textbf{4.085} & 4.502 & \textbf{4.002} & \textbf{3.114} & \textbf{4.923} & \textbf{4.154} \\
w/o InfoNCE                 & 4.302 & 4.072 & 4.481 & 3.964 & 3.078 & 4.921 & 4.136 \\
w/o entity-separated        & 4.284 & 4.053 & \textbf{4.508} & 3.963 & 3.052 & 4.916 & 4.129 \\
w/o saliency head           & 4.222 & 3.976 & 4.436 & 3.917 & 2.998 & 4.916 & 4.078 \\
w/o temporal mask           & \textbf{4.312} & 4.073 & 4.500 & 3.952 & 3.091 & 4.915 & 4.140 \\
XOR router                  & 4.292 & 4.016 & 4.470 & 3.931 & 3.035 & 4.921 & 4.111 \\
w/ temporal decay $\tau=10$ & 4.310 & 4.066 & 4.496 & 3.982 & 3.062 & 4.915 & 4.139 \\
\bottomrule
\end{tabular}
\end{table}

\begin{table}[htbp]
\centering
\caption{\textbf{Per-sub-dimension motion-quality scores (mean across three VLM judges).} Best non-oracle in \textbf{bold}. \emph{MtPrm}=motion\_prompt\_alignment, \emph{MtCmp}=motion\_completeness, \emph{MtAmp}=motion\_amplitude, \emph{TmpCns}=temporal\_consistency, \emph{PhyPlu}=physical\_plausibility, \emph{Cam}=camera\_motion, \emph{Inter}=interaction\_correctness.}
\label{tab:appendix:motion-dims}
\footnotesize
\begin{tabular}{lcccccccc}
\toprule
Method & MtPrm & MtCmp & MtAmp & TmpCns & PhyPlu & Cam & Inter & Avg \\
\midrule
\multicolumn{9}{l}{\emph{Main comparison}} \\
\midrule
Real video (oracle)         & 3.950 & 4.289 & 4.485 & 4.561 & 4.636 & 4.828 & 4.272 & 4.431 \\
\midrule
Pretrained Wan2.2           & 2.502 & 3.338 & 3.765 & \textbf{4.610} & \textbf{4.654} & 4.608 & 3.250 & 3.818 \\
SFT                         & 2.651 & 3.292 & 3.700 & 4.405 & 4.509 & 4.577 & 3.355 & 3.784 \\
VideoREPA                   & 2.676 & 3.306 & 3.713 & 4.420 & 4.522 & 4.608 & 3.372 & 3.802 \\
MoAlign                     & 2.689 & 3.332 & 3.742 & 4.379 & 4.493 & 4.594 & 3.381 & 3.802 \\
\textbf{SARA (ours)}        & \textbf{2.744} & \textbf{3.396} & \textbf{3.807} & 4.431 & 4.537 & \textbf{4.623} & \textbf{3.423} & \textbf{3.852} \\
\midrule
\multicolumn{9}{l}{\emph{Ablations}} \\
\midrule
\textbf{SARA (full)}        & \textbf{2.744} & \textbf{3.396} & \textbf{3.807} & 4.431 & 4.537 & \textbf{4.623} & \textbf{3.423} & \textbf{3.852} \\
w/o InfoNCE                 & 2.676 & 3.336 & 3.762 & 4.378 & 4.477 & 4.617 & 3.380 & 3.804 \\
w/o entity-separated        & 2.673 & 3.328 & 3.722 & 4.420 & 4.519 & 4.618 & 3.390 & 3.810 \\
w/o saliency head           & 2.621 & 3.265 & 3.681 & \textbf{4.439} & \textbf{4.554} & 4.610 & 3.326 & 3.785 \\
w/o temporal mask           & 2.698 & 3.369 & 3.802 & 4.389 & 4.485 & 4.585 & 3.379 & 3.815 \\
XOR router                  & 2.663 & 3.327 & 3.746 & 4.414 & 4.512 & 4.593 & 3.375 & 3.804 \\
w/ temporal decay $\tau=10$ & 2.698 & 3.373 & 3.802 & 4.432 & 4.528 & 4.611 & 3.377 & 3.832 \\
\bottomrule
\end{tabular}
\end{table}

\section{Detailed VBench results}
\label{sec:appendix:more_results}

This section reports the per-task scores underlying the dimension-level VBench-1.0~\citep{huang2024vbench} and VBench-2.0~\citep{zheng2025vbench} entries of Tab.~\ref{tab:vbench_combined} (Sec.~\ref{sec:exp:vbench}).

\subsection{VBench-1.0 per-task semantic scores}
\label{sec:appendix:vbench1}

VBench-1.0 splits its $16$ atomic tasks into a \emph{Quality} dimension and a \emph{Semantic} dimension. We evaluate the five continually-trained Wan2.2 high-noise checkpoints on the nine atomic tasks that compose the \emph{Semantic} dimension: \emph{Scene}, \emph{Overall Consistency}, \emph{Appearance Style}, \emph{Object Class}, \emph{Spatial Relationship}, \emph{Human Action}, \emph{Temporal Style}, \emph{Color}, and \emph{Multiple Objects}. The aggregate \emph{Semantic} column is the official mean over these nine tasks, after the per-task normalisation defined in the VBench-1.0 release. Tab.~\ref{tab:appendix:vbench1-tasks} reports the raw per-task scores, and for readability we keep the official $0$--$1$ range rather than rescaling to $\%$.

\begin{table}[htbp]
\centering
\caption{\textbf{VBench-1.0 per-task semantic scores} (raw, $0$--$1$ range as returned by the official scorers, higher is better; aggregated over the official $946$-prompt suite). The nine columns are the official VBench-1.0 \emph{Semantic} sub-tasks, and the rightmost \emph{Semantic (Avg.)} column is the official semantic-dimension aggregate (the percentage version of this column is repeated under \emph{VBench-1.0 / Semantic} in Tab.~\ref{tab:vbench_combined}). Best per column in \textbf{bold}.}
\label{tab:appendix:vbench1-tasks}
\small
\setlength{\tabcolsep}{3.5pt}
\resizebox{\linewidth}{!}{%
\begin{tabular}{l|ccccccccc|c}
\toprule
Method & Scene & Consistency & Appearance & Object & Spatial & Action & Temporal & Color & Multiple & Semantic (Avg.) \\
\midrule
Pretrained Wan2.2                            & 0.3401 & \textbf{0.2524} & 0.2101 & 0.8560 & 0.7631 & 0.8800 & 0.2315 & 0.9012 & 0.6677 & 0.7274 \\
SFT                                          & 0.3481 & 0.2436 & 0.2048 & 0.8449 & 0.8074 & 0.8100 & 0.2187 & 0.9100 & 0.7134 & 0.7217 \\
VideoREPA~\citep{zhang2025videorepa}         & 0.2943 & 0.2511 & \textbf{0.2125} & \textbf{0.8829} & 0.7699 & \textbf{0.8900} & \textbf{0.2319} & 0.8957 & 0.7027 & 0.7299 \\
MoAlign~\citep{bhowmik2025moalign}           & \textbf{0.3583} & 0.2476 & 0.2105 & 0.8275 & \textbf{0.8108} & 0.8500 & 0.2232 & 0.8870 & 0.7248 & 0.7295 \\
\textbf{SARA (ours)}                         & \textbf{0.3583} & 0.2487 & 0.2071 & 0.8758 & 0.7710 & 0.8400 & 0.2298 & \textbf{0.9313} & \textbf{0.7576} & \textbf{0.7389} \\
\bottomrule
\end{tabular}%
}
\end{table}

\paragraph{Per-task semantic scores.} The per-task picture (Tab.~\ref{tab:appendix:vbench1-tasks}) is diffuse, as expected for nine sub-tasks covering very different aspects of text-following: SARA leads the multi-entity-heavy tasks it targets (\emph{Multiple Objects}, \emph{Color}) while the remaining tasks split across baselines, and the pretrained model's small lead on \emph{Overall Consistency} is the VBench-1.0 instance of the high-noise-only-training trade-off discussed in Sec.~\ref{sec:exp:vbench} and App.~\ref{sec:appendix:limitations}. SARA nonetheless cleanly leads the aggregate \emph{Semantic} score of Tab.~\ref{tab:vbench_combined}.

\subsection{VBench-2.0 per-task scores}
\label{sec:appendix:vbench2}

VBench-2.0~\citep{zheng2025vbench} groups its $18$ atomic tasks into five dimensions: \emph{creativity} (Composition, Diversity), \emph{commonsense} (Instance Preservation, Motion Rationality), \emph{controllability} (Camera Motion, Complex Landscape, Complex Plot, Dynamic Attribute, Dynamic Spatial Relationship, Human Interaction, Motion Order Understanding), \emph{human fidelity} (Human Anatomy, Human Clothes, Human Identity), and \emph{physics} (Material, Mechanics, Multi-View Consistency, Thermotics). Each dimension score is the mean of its tasks, and the final score is the mean of the five dimensions. Tab.~\ref{tab:appendix:vbench2-tasks} reports all $18$ per-task scores plus the VBench-2.0 final score.

\begin{table}[htbp]
\centering
\caption{\textbf{VBench-2.0 per-task scores under all five dimensions} ($\%$, higher is better; aggregated over the official $1{,}013$-prompt suite, $3$ generations per prompt and $20$ for the special \emph{Diversity} task). The $19$ columns are wrapped into three horizontal stripes that share a single \textbf{Method} header column: stripe 1 covers \emph{Creativity} + \emph{Commonsense} + \emph{Human Fidelity}, stripe 2 covers \emph{Controllability}, and stripe 3 covers \emph{Physics} together with the overall VBench-2.0 \emph{Final} score (mean of the five dimensions, repeated from Tab.~\ref{tab:vbench_combined}). Best per column in \textbf{bold}.}
\label{tab:appendix:vbench2-tasks}
\scriptsize
\setlength{\tabcolsep}{4pt}
\renewcommand{\arraystretch}{1.1}
\resizebox{\linewidth}{!}{%
\begin{tabular}{l|ccccccc}
\toprule
\textbf{Method} & \textbf{Composition} & \textbf{Diversity} & \textbf{Instance Preservation} & \textbf{Motion Rationality} & \textbf{Human Anatomy} & \textbf{Human Clothes} & \textbf{Human Identity} \\
\midrule
Pretrained Wan2.2    & 45.69          & 59.43          & 86.55          & 30.46          & \textbf{90.56} & \textbf{86.67} & \textbf{80.88} \\
SFT                  & 45.79          & 63.42          & \textbf{90.06} & 29.31          & 88.06          & 77.90          & 75.27 \\
VideoREPA            & 48.09          & 60.06          & 88.89          & \textbf{33.33} & 87.86          & 84.23          & 76.26 \\
MoAlign              & 49.48          & \textbf{64.01} & 88.89          & 30.46          & 89.26          & 84.97          & 80.02 \\
\textbf{SARA (ours)} & \textbf{50.81} & 59.95          & 88.89          & \textbf{33.33} & 89.05          & 85.78          & 80.38 \\
\midrule
\textbf{Method} & \textbf{Camera Motion} & \textbf{Complex Landscape} & \textbf{Complex Plot} & \textbf{Dynamic Attribute} & \textbf{Dynamic Spatial Relationship} & \textbf{Human Interaction} & \textbf{Motion Order Understanding} \\
\midrule
Pretrained Wan2.2    & 15.79          & 18.44          & 11.56          & 38.46          & \textbf{41.55} & 63.67          & \textbf{27.36} \\
SFT                  & 16.36          & 16.89          & \textbf{12.76} & 41.39          & 40.10          & 58.00          & 21.62 \\
VideoREPA            & 17.90          & 20.00          & 10.67          & 44.69          & 36.23          & \textbf{65.00} & 26.26 \\
MoAlign              & 17.90          & \textbf{20.44} & 12.67          & 41.39          & 34.30          & 62.33          & 21.55 \\
\textbf{SARA (ours)} & \textbf{19.14} & 15.78          & 11.07          & \textbf{46.89} & 37.20          & 59.67          & 26.60 \\
\midrule
\textbf{Method} & \textbf{Material} & \textbf{Mechanics} & \textbf{Multi-View Consistency} & \textbf{Thermotics} & \textbf{Final} & & \\
\midrule
Pretrained Wan2.2    & 43.24          & \textbf{53.54} & 38.99          & 51.80          & 55.00          & & \\
SFT                  & \textbf{54.17} & 47.62          & \textbf{45.65} & \textbf{56.92} & 55.08          & & \\
VideoREPA            & 44.59          & 47.45          & 40.57          & 54.07          & 55.24          & & \\
MoAlign              & 49.30          & 46.88          & 42.11          & 52.99          & 55.81          & & \\
\textbf{SARA (ours)} & 47.89          & 48.51          & 44.17          & 53.44          & \textbf{56.19} & & \\
\bottomrule
\end{tabular}%
}
\end{table}

\paragraph{Per-task scores.} The $18$ per-task scores (Tab.~\ref{tab:appendix:vbench2-tasks}) are likewise diffuse: SARA concentrates its wins on the multi-entity-composition and dynamic-attribute tasks that saliency routing targets (e.g.\ \emph{Composition}, \emph{Dynamic Attribute}, \emph{Camera Motion}), while the baselines split the remaining tasks and the pretrained model retains the expected small lead on the Human-Fidelity and other low-noise-rendered tasks (the high-noise-only-training trade-off of App.~\ref{sec:appendix:limitations}, on which SARA shows the smallest drop). SARA still leads the aggregate VBench-2.0 \emph{Final} score of Tab.~\ref{tab:vbench_combined}.

\section{Additional ablation: DiT alignment hookup layer}
\label{sec:appendix:layer_ablation}
\label{sec:appendix:layer_ablation:hookup}

The Stage~2 masked TRD loss of Eq.~\eqref{eq:lmtrd} is computed on the projected hidden state $V_p$ of a single Wan2.2 DiT layer. The high-noise DiT has $40$ layers, and the main paper hooks the loss into layer $18$ (mid-depth, Tab.~\ref{tab:appendix:hparams-stage2}). Tab.~\ref{tab:appendix:layer-ablation} moves the hookup to deeper layers ($30$, $36$, $39$) with all else fixed at the default SARA configuration. Layer $18$ is the published optimum of prior REPA-family work on shallower DiTs (VideoREPA on CogVideoX~\citep{zhang2025videorepa}, MoAlign on Wan2.1~\citep{bhowmik2025moalign}); the open question is whether Wan2.2's larger depth budget shifts the optimum deeper, and Tab.~\ref{tab:appendix:layer-ablation} shows it does not. We do not re-test shallower hookups, since REPA's sweep on DiT-XL/2~\citep{yu2024representation} found pre-mid-depth blocks carry mostly low-level and positional signal.

Layer $18$ dominates on all four VLM-rubric metrics, with every deeper hookup behind it by up to $0.055$ (TA mean) and $0.039$ (MQ mean). Differences among the deeper hookups are small ($\leq 0.04$) and non-monotonic in layer index, so the operative distinction is mid- vs late-depth rather than the precise late-layer position. A plausible reason is that mid-depth blocks still carry spatially localised but semantically structured tokens, whereas the latest layers specialise toward noise prediction and align less well with the V-JEPA target. We therefore use layer $18$ throughout.

\begin{table}[htbp]
\centering
\caption{\textbf{Effect of the DiT alignment hookup layer on Wan2.2 high-noise} ($40$ DiT layers total). Each row retrains SARA with the masked TRD loss of Eq.~\eqref{eq:lmtrd} attached to a different layer, with all other settings matching the default SARA configuration. Metrics follow the VLM rubric of Tab.~\ref{tab:main}. Best in \textbf{bold}.}
\label{tab:appendix:layer-ablation}
\small
\begin{tabular}{lcccc}
\toprule
Hookup layer & TA mean & TA vote & MQ mean & MQ vote \\
\midrule
\textbf{layer $18$ (default, mid-depth)} & \textbf{4.1543} & \textbf{4.1668} & \textbf{3.8516} & \textbf{3.9191} \\
layer $30$                               & 4.0990 & 4.1221 & 3.8266 & 3.8972 \\
layer $36$                               & 4.1214 & 4.1404 & 3.8129 & 3.8797 \\
layer $39$                               & 4.1368 & 4.1569 & 3.8354 & 3.8946 \\
\bottomrule
\end{tabular}
\end{table}

\section{Training details}
\label{sec:appendix:training}

Tables~\ref{tab:appendix:hparams-stage1} and~\ref{tab:appendix:hparams-stage2} list all training-side hyperparameters for SARA's two stages. The Stage~2 configuration applies verbatim to SFT (auxiliary loss disabled), VideoREPA (no saliency routing), and the MoAlign reproduction (motion subspace $D_m = 64$, projector $P_\zeta$, exponential temporal decay $\tau{=}10$, matching \citet{bhowmik2025moalign}), and only the auxiliary objective changes. Stage~2 follows Wan2.2's two-stage timestep partition (boundary ratio $0.875$, visible in Tab.~\ref{tab:appendix:hparams-stage2}) and trains only the high-noise transformer.

\begin{table}[htbp]
\centering
\caption{\textbf{Stage 1 training hyperparameters: saliency aligner.}}
\label{tab:appendix:hparams-stage1}
\footnotesize
\begin{tabular}{ll}
\toprule
\textbf{Setting} & \textbf{Value} \\
\midrule
\multicolumn{2}{l}{\emph{Data}} \\
\midrule
Training corpus              & $500$K MTSS-recaptioned clips \\
Frames per clip $T$          & $32$ \\
Input resolution             & dynamic, max edge $480$ \\
V-JEPA encoder input         & $384$ (ViT-G/16, patch $16$, tubelet $2$) \\
Entity caption length cap    & $2048$ tokens \\
\midrule
\multicolumn{2}{l}{\emph{Model}} \\
\midrule
Trainable modules            & $\Phi_\text{CA} + \Phi_\text{sal} + \Phi_\text{proj}$ \\
Frozen modules               & V-JEPA 2.1 ViT-G/16, SAM 3.1 Multiplex, Qwen3-VL-Emb-2B \\
V-JEPA input / patch / tubelet & $384$ / $16$ / $2$ \\
Visual dim $D_v$             & $1664$ \\
LM input-embed dim $D_t$     & Qwen3-VL-Emb-2B native \\
$\Phi_\text{CA}$ architecture            & $6$ blocks ($2$ CA, $4$ SA), $8$ heads, no pos.\ emb. \\
$\Phi_\text{sal}$ architecture           & 2-layer MLP $\mathbb{R}^{D_v}{\to}\mathbb{R}^{512}{\to}\mathbb{R}$, sigmoid \\
$\Phi_\text{proj}$ architecture          & RMSNorm $+$ linear $\mathbb{R}^{D_v}{\to}\mathbb{R}^{D_t}$ \\
InfoNCE temperature $\tau_\text{nce}$   & $0.07$ \\
\midrule
\multicolumn{2}{l}{\emph{Loss}} \\
\midrule
Objective                    & $\lambda_\text{BCE}\mathcal{L}_\text{BCE}+\lambda_\text{InfoNCE}\mathcal{L}_\text{InfoNCE}$ \\
Loss weights                 & $\lambda_\text{BCE} = \lambda_\text{InfoNCE} = 1$ \\
Supervision units            & $K$ per-entity $(c_k, M_k)$ + $(c_\text{fg}, M_\text{fg})$ + $(c_\text{bg}, M_\text{bg})$ \\
\midrule
\multicolumn{2}{l}{\emph{Optimization}} \\
\midrule
Optimizer                    & AdamW \\
Peak / min learning rate     & $5 \times 10^{-5}$ / $1 \times 10^{-6}$ (cosine schedule) \\
LR warmup steps              & $500$ (linear) \\
Weight decay                 & $0.01$ \\
Gradient clipping            & $1.0$ \\
Per-GPU batch size           & $2$ \\
Gradient accumulation        & $1$ \\
Mixed precision              & bf16 \\
\midrule
\multicolumn{2}{l}{\emph{Distributed setup}} \\
\midrule
Number of GPUs               & $32$ \\
Effective batch (Reference-stream forwards) & $64$ \\
Total training steps         & $3{,}000$ \\
Gradient checkpointing       & enabled \\
\bottomrule
\end{tabular}
\end{table}

\begin{table}[htbp]
\centering
\caption{\textbf{Stage 2 training hyperparameters: diffusion continual training.}}
\label{tab:appendix:hparams-stage2}
\footnotesize
\begin{tabular}{ll}
\toprule
\textbf{Setting} & \textbf{Value} \\
\midrule
\multicolumn{2}{l}{\emph{Data}} \\
\midrule
Training corpus              & $500$K MTSS-recaptioned clips \\
Frames per clip $T$          & $81$ \\
Input resolution             & $480{\times}848$ \\
V-JEPA encoder input         & $384$ (ViT-G/16, patch $16$, tubelet $2$) \\
\midrule
\multicolumn{2}{l}{\emph{Model}} \\
\midrule
Trainable modules            & Wan2.2 high-noise DiT ($\sim$$14$B) \\
Frozen modules               & V-JEPA 2.1, Stage 1 aligner, Wan2.2 VAE, low-noise DiT \\
REPA target dim              & $1664$ \\
DiT alignment hookup         & layer $18$ \\
\midrule
\multicolumn{2}{l}{\emph{Loss}} \\
\midrule
Objective                    & $\mathcal{L}_\text{diff}+\lambda_\text{TRD}\,\mathcal{L}_\text{m\text{-}TRD}$ \\
Loss weight $\lambda_\text{TRD}$         & $0.5$ \\
Spatial / temporal balance $\lambda_\text{tmp}$ & $1.0$ \\
Numerical floor $\varepsilon$ & $10^{-6}$ \\
Pair-routing operator        & OR ($w_i + w_j - w_iw_j$) \\
Saliency mask scope          & spatial $+$ temporal \\
Temporal decay $\tau$        & $\infty$ (uniform cross-frame) \\
\midrule
\multicolumn{2}{l}{\emph{Optimization}} \\
\midrule
Optimizer                    & AdamW \\
Learning rate                & $5 \times 10^{-6}$ (constant after warmup) \\
LR warmup steps              & $500$ (linear) \\
Weight decay                 & $0.01$ \\
Gradient clipping            & $1.0$ \\
Per-GPU batch size           & $1$ \\
Gradient accumulation        & $1$ \\
Mixed precision              & bf16 \\
Timestep weighting           & logit-normal, sample shift $12.0$ \\
Timestep range               & $t \in [0.875,\,1.0] \cdot 1000$ (high-noise) \\
Random seed                  & $1024$ \\
\midrule
\multicolumn{2}{l}{\emph{Distributed setup}} \\
\midrule
Sequence-parallel size       & $1$  \\
Effective batch (videos)     & $40$ \\
Total training steps         & $3{,}000$ \\
\bottomrule
\end{tabular}
\end{table}

\section{Limitations and broader impact}
\label{sec:appendix:limitations}

\paragraph{Caption-pipeline noise.} Stage~1 SAM masks and Stage~2 conditioning both consume MTSS captions produced by a frozen Qwen3-VL-72B rewriter (App.~\ref{sec:appendix:mtss}), which can occasionally hallucinate entities or mis-bind attributes. Because the same captions are used by every row of Table~\ref{tab:main} (pretrained, SFT, VideoREPA, MoAlign, SARA), this noise enters as a constant offset that does not affect the SARA-vs-baseline ranking. It shows up only as a sub-$5$ ceiling on the \emph{Real video} oracle row (App.~\ref{sec:appendix:vlmrubric}). The same ranking holds under the strict-intersection judge filter, the rubric-independent user study (Sec.~\ref{sec:exp:userstudy}), and the VBench protocols (Sec.~\ref{sec:exp:vbench}), so the SARA gains are not an artefact of caption-rewriter noise.

\paragraph{High-noise-only continual training induces a small low-noise distribution shift.} Wan2.2 ships as a two-expert MoE: a high-noise expert that handles the early, coarse-structure stages of denoising and a low-noise expert that renders fine details (human anatomy and identity, fine textures, and high-frequency temporal structure) at late, low-noise timesteps. Our matched protocol, shared by every continually-trained row in Tab.~\ref{tab:main} (SFT, VideoREPA, MoAlign, SARA), continually trains only the high-noise expert and leaves the low-noise expert frozen. Any continual-training method that improves coarse-structure prompt following therefore shifts the intermediate-latent distribution that the un-updated low-noise expert was trained against, producing a small but consistent drop on the low-noise-rendered dimensions: VBench-2.0 \emph{Human Fidelity} (Human Anatomy, Human Clothes, Human Identity; Sec.~\ref{sec:exp:vbench}, App.~\ref{sec:appendix:vbench2}), VBench-1.0 \emph{Overall Consistency} (App.~\ref{sec:appendix:vbench1}), and the VLM-rubric \texttt{temporal\_consistency} / \texttt{physical\_plausibility} sub-dimensions (App.~\ref{sec:appendix:vlmrubric}). The drop appears on all continually-trained methods, and SARA shows the smallest one, consistent with text-conditioned saliency producing the most targeted high-noise update and therefore the smallest distribution shift on the low-noise expert. The clean fix, jointly training both experts under the same SARA objective, is left to future work, and the matched-setting comparison is unaffected because the same single-expert constraint applies to every row.

\paragraph{Broader impact.} SARA does not introduce new generative capability, and instead reallocates an existing alignment loss on top of an already-released VDM. It therefore inherits, rather than amplifies, the standard text-to-video dual-use risks (deepfakes, biased depictions, copyrighted-style imitation), and any deployment should keep the safety-tuning, watermarking, and content-filtering layers that ship with the base-model release.

\bigskip
\label{references}
\addcontentsline{toc}{section}{References}
\bibliographystyle{assets/plainnat}
\bibliography{paper}

\end{document}